\documentclass[acmsmall]{acmart}
\usepackage{multirow}
\usepackage{graphics}
\usepackage{color}
\usepackage{amsthm,amsmath}
\usepackage{amssymb}
\AtBeginDocument{%
  \providecommand\BibTeX{{%
    \normalfont B\kern-0.5em{\scshape i\kern-0.25em b}\kern-0.8em\TeX}}}

\setcopyright{acmcopyright}
\acmJournal{TOIS}
\acmYear{2021} \acmVolume{1} \acmNumber{1} \acmArticle{1} \acmMonth{1} \acmPrice{15.00}\acmDOI{10.1145/3464377}

\begin{document}

\title{Unstructured Text Enhanced Open-domain Dialogue System: A Systematic Survey}

\author{Longxuan Ma}
\email{lxma@ir.hit.edu.cn}
\orcid{0000-0003-1431-8485}
\affiliation{%
  \institution{Research Center for Social Computing and Information Retrieval, Harbin Institute of Technology}
  \streetaddress{92, Xidazhi Road, Nangang Qu}
  \city{Harbin}
  \state{Heilongjiang}
  \country{China}
}

\author{Mingda Li}
\email{mdli@ir.hit.edu.cn}
\orcid{}
\affiliation{%
  \institution{Research Center for Social Computing and Information Retrieval, Harbin Institute of Technology}
  \streetaddress{92, Xidazhi Road, Nangang Qu}
  \city{Harbin}
  \state{Heilongjiang}
  \country{China}
}

\author{Wei-Nan Zhang}
\authornote{Corresponding author.}
\email{wnzhang@ir.hit.edu.cn}
\orcid{0000-0001-5553-178X}
\affiliation{  
  \institution{Research Center for Social Computing and Information Retrieval, Harbin Institute of Technology}
  \streetaddress{92, Xidazhi Road, Nangang Qu}
  \city{Harbin}
  \state{Heilongjiang}
  \country{China}
}

\author{Jiapeng Li}
\email{jpli@ir.hit.edu.cn}
\orcid{}
\affiliation{  
  \institution{Research Center for Social Computing and Information Retrieval, Harbin Institute of Technology}
  \streetaddress{92, Xidazhi Road, Nangang Qu}
  \city{Harbin}
  \state{Heilongjiang}
  \country{China}
}

\author{Ting Liu}
\email{tliu@ir.hit.edu.cn}
\orcid{}
\affiliation{%
  \institution{Research Center for Social Computing and Information Retrieval, Harbin Institute of Technology}
  \streetaddress{92, Xidazhi Road, Nangang Qu}
  \city{Harbin}
  \state{Heilongjiang}
  \country{China}
}
\begin{abstract}
Incorporating external knowledge into dialogue generation has been proven to benefit the performance of an open-domain Dialogue System (DS), such as generating informative or stylized responses, controlling conversation topics. In this article, we study the open-domain DS that uses unstructured text as external knowledge sources (\textbf{U}nstructured \textbf{T}ext \textbf{E}nhanced \textbf{D}ialogue \textbf{S}ystem, \textbf{UTEDS}). The existence of unstructured text entails distinctions between UTEDS and traditional data-driven DS and we aim to analyze these differences. We first give the definition of the UTEDS related concepts, then summarize the recently released datasets and models. We categorize UTEDS into Retrieval and Generative models and introduce them from the perspective of model components. The retrieval models consist of Fusion, Matching, and Ranking modules, while the generative models comprise Dialogue and Knowledge Encoding, Knowledge Selection, and Response Generation modules. We further summarize the evaluation methods utilized in UTEDS and analyze the current models' performance. At last, we discuss the future development trends of UTEDS, hoping to inspire new research in this field. 
\end{abstract}




\maketitle

\section{Introduction}

%%======old history to task-oriented%%=====work try to combine open-domain and task-oriented, then into open-domain
Dialogue systems (DS) attract great attention because of its wide application prospects. Early DS such as Eliza \cite{DBLP:journals/cacm/Weizenbaum66}, Parry \cite{DBLP:journals/ai/ColbyWH71}, and Alice \cite{DBLP:books/daglib/0072558} attempted to imitate human behaviors in conversations and challenge different forms of the Turing Test \cite{turing2009computing}. They worked well but only in constrained environments, an open-domain DS remained an elusive task until recently \cite{DBLP:journals/corr/abs-1905-05709}. 

%introducing external knowledge and compare structure and unstructured

One of the main challenges in open-domain DS is that the generated responses lack sufficient information \cite{DBLP:conf/naacl/LiGBGD16}. Previous researchers proposed different methods to alleviate this issue. Diversity enhancement strategies \cite{DBLP:conf/acl/ZhaoZE17,DBLP:conf/iclr/HoltzmanBDFC20,DBLP:conf/naacl/GaoLZBGGD19,DBLP:conf/iclr/WelleckKRDCW20} and large scale parameters \cite{radford2019language,DBLP:conf/acl/ZhangSGCBGGLD20,DBLP:journals/corr/abs-2005-14165} have been proven to be effective. In addition to the above methods, introducing external knowledge, such as structured knowledge graph (KG) \cite{DBLP:conf/emnlp/ParthasarathiP18,DBLP:conf/emnlp/AgarwalDKR18,DBLP:conf/emnlp/Wu00W20,DBLP:conf/aaai/0006WHWL019} or unstructured text \cite{DBLP:conf/aaai/GhazvininejadBC18,DBLP:conf/emnlp/ParthasarathiP18,DBLP:conf/acl/KielaWZDUS18,DBLP:journals/corr/abs-1811-01241,DBLP:conf/emnlp/ZhouPB18,DBLP:conf/sigir/YangQQGZCHC18}, into the dialogue generation attracts great attention. \textit{External knowledge acts as an extra resource besides the context of dialogues and provides more diversified and grounded information for dialogue generation.}

%?????????should knowledge base include here?????????
Structured knowledge usually means \textbf{Knowledge Graph}\footnote{Following \citet{DBLP:journals/corr/abs-2002-00388}, we use the terms knowledge graph and knowledge base interchangeably in this paper.}, which is a semantic map obtained after defining and extracting the named entities and the relations \cite{DBLP:conf/ijcai/ZhouYHZXZ18,DBLP:conf/cikm/0005HQQGCLSL19} in unstructured text. It was first proposed to better understand the natural language used in search engine\footnote{https://www.blog.google/products/search/introducing-knowledge-graph-things-not-strings}. Although it has been widely applicated in natural language processing (NLP) \cite{DBLP:conf/kdd/0001GHHLMSSZ14,DBLP:journals/pieee/Nickel0TG16,DBLP:journals/tkde/WangMWG17,DBLP:journals/corr/abs-2003-02320} and achieved great success, some researchers pointed out that complex reasoning \cite{DBLP:conf/nips/Qu019,DBLP:conf/iclr/ZhangCYRLQS20}, unified representation framework \cite{DBLP:conf/icml/LiuWY17,DBLP:conf/acl/HayashiS17,DBLP:conf/aaai/Han0S18}, interpretability \cite{DBLP:conf/acl/XieMDH17,DBLP:conf/wsdm/ZhangPZBC19}, scalability \cite{DBLP:conf/aaai/NickelRP16,DBLP:conf/nips/YangYC17}, knowledge aggregation \cite{DBLP:conf/emnlp/PetroniRRLBWM19}, and knowledge automatic extraction \cite{DBLP:journals/corr/abs-2002-00388} were still facing challenges. Besides the traditional KG, there are other forms of knowledge, to some extent, that can be seen as structured and used as external knowledge. For example, \textbf{Event Graph} \cite{DBLP:conf/ijcai/XuLWN0C20} or \textbf{Event Logic Graph} \cite{DBLP:journals/corr/abs-1907-08015} uses event descriptions (a sentence or a predicate phrase) as vertices and logical relations (causal, sequential, etc.) between events as edges; \textbf{Enhanced KG} \cite{DBLP:conf/acl/ZhouZHHZ20,DBLP:conf/acl/LinJHWC20} takes a KG as its backbone and aligns unstructured text to related entities; \citet{DBLP:conf/emnlp/ChenSYW20} used \textbf{Table Text} \cite{DBLP:conf/emnlp/LebretGA16} as external knowledge for a data-to-text generation task.

\begin{table*}[h]
\footnotesize
  \caption{Classification according to the properties of text.}
   \label{Classification}
   \begin{tabular}{l|l}
    \toprule
    Category & Examples \\
    \midrule
    \textbf{Factual description}  & Wikipedia articles \cite{DBLP:journals/corr/abs-1811-01241,DBLP:conf/acl/QinGBLGDCG19}, News reports \cite{DBLP:conf/interspeech/GopalakrishnanH19,DBLP:journals/corr/abs-2005-12529}, Domain specific knowledge \cite{DBLP:conf/www/SzpektorCEFHKKO20}\\

\midrule
\textbf{Fictional information}  & Stories \cite{DBLP:conf/ijcai/XuLWN0C20}, Novel \cite{DBLP:conf/emnlp/GaoZLGBGD19}, Persona information \cite{DBLP:conf/acl/KielaWZDUS18}, Virtual world description \cite{DBLP:conf/emnlp/UrbanekFKJHDRKS19}.\\
& Movie script \cite{DBLP:conf/www/TigunovaYMW19}, Movie plots \cite{DBLP:conf/emnlp/ZhouPB18}, TV dialogues \cite{DBLP:conf/interspeech/SuHTL19}, Emotions \cite{DBLP:conf/acl/RashkinSLB19}.\\
\midrule
\textbf{Subjective comments} & Reviews of goods \cite{DBLP:conf/aaai/GhazvininejadBC18} or movies \cite{DBLP:conf/emnlp/MogheABK18}, Reddit comments \cite{DBLP:journals/corr/abs-1901-03461}, Interview \cite{DBLP:conf/emnlp/MajumderLNM20} \\%Dialogues \cite{DBLP:conf/emnlp/WestonDM18,DBLP:conf/aaai/0006WHWL019} 
\bottomrule
  \end{tabular}
\end{table*}

%add more details about each advantage
%concrete knowledge and abstractive knowledge
Unstructured text can be \textbf{Factual description}, \textbf{Fictional information}, or \textbf{Subjective comments}, as shown in Table \ref{Classification}. Factual descriptions are statements of objective facts such as news reports \cite{DBLP:conf/interspeech/GopalakrishnanH19}, historical events \cite{DBLP:journals/corr/abs-1811-01241} and domain specific knowledge \cite{DBLP:conf/www/SzpektorCEFHKKO20}. Fictional information is usually artificially constructed text such as stories \cite{DBLP:conf/ijcai/XuLWN0C20}, movie script \cite{DBLP:conf/www/TigunovaYMW19} or TV dialogues \cite{DBLP:conf/interspeech/SuHTL19}. Subjective comments are usually reviews of things in the world that are recognized by more people on social networks \cite{DBLP:conf/aaai/GhazvininejadBC18,DBLP:conf/emnlp/MogheABK18}. Generally, comments with a high degree of recognition have high authenticity.

%%%should introduce the difficulty of using unstructured text and the facilitate of using structured knowledge

Apart from the categories above, unstructured text can also be divided into \textbf{independent sentences} and \textbf{documents}. Independent sentences usually do not have logical relationships between themselves \cite{DBLP:conf/coling/VougiouklisHS16,DBLP:conf/aaai/GhazvininejadBC18,DBLP:conf/acl/KielaWZDUS18}. In contrast, a document contains multiple logically related sentences, which together constitute a description of the topic of the document \cite{DBLP:conf/emnlp/ZhouPB18,DBLP:conf/emnlp/MogheABK18,DBLP:conf/acl/QinGBLGDCG19}. But in the current stage, most works treat them in a similar way. Hence in this paper, we use a unified term "unstructured text" to integrate them. But we do believe that there are more sophisticated semantic structures that can be extracted from documents in future work. Table \ref{SentDoc} presents examples of independent sentences and a document in UTED tasks. The dialogue using document as external knowledge is also called document-grounded conversation (DGC) \cite{DBLP:conf/emnlp/ZhouPB18} or background-based conversation (BBC) \cite{DBLP:conf/sigir/MengRCSRTR20}. Notably, the Conversational QA is also based on background documents. The differences between DGC/BBC and Conversational QA (such as CoQA \cite{DBLP:journals/tacl/ReddyCM19}, QuAC \cite{DBLP:conf/emnlp/ChoiHIYYCLZ18} and MANTIS \cite{DBLP:journals/corr/abs-1912-04639}) are the dialogue of DGC/BBC is more diversified (including chit-chat or recommendation) and not limited to QA. 

%We do not cover conversational QA in this paper since DGC/BBC 
%is not limited to information-seeking, but also includes chit-chat, \footnote{The DGC/BBC task not only select the related information (word, segment, sentence, etc.) in the background, but also use these background information in the right place during decoding.}. %Besides, it is usually treated as a generation task instead of span extraction \cite{DBLP:journals/tacl/ReddyCM19} task. 

%\texttt{} show equal distance
%\underline can't change line, \uline can
%\usepackage{ulem} can get a better underline, other effects are:
%\uuline{double underline}\par
%\uwave{wave line}\par
%\sout{delete line in the middle}\par
%\xout{delete line iteit}\par
%\dashuline{dash underline}\par
%\dotuline{dot underline}
\begin{table*}
\footnotesize
  \caption{The examples of independent sentences and a document as external knowledge in UTED tasks. The information used in the dialogue is marked with the same color and font in the external text.}
   \label{SentDoc}
   \begin{tabular}{l}
    \toprule
\qquad\qquad\qquad\qquad\qquad\qquad Sentences as external knowledge (Persona-Chat \cite{DBLP:conf/acl/KielaWZDUS18}) \\
\midrule
"I am an artist"; "\textcolor{red}{\textit{I have four children}}"; "I recently got a cat";"I enjoy walking ..."; "I love \textcolor{blue}{\textbf{watching Game of Thrones}}"\\
\midrule
Speaker 1: \textcolor{red}{\textit{My children}} and I were just about to \textcolor{blue}{\textbf{watch Game of Thrones}}. \\
Speaker 2: Nice! How old are your children? \\
Speaker 1: \textcolor{red}{\textit{I have four}} that range in age from 10 to 21. You? \\
Speaker 2: I do not have children at the moment.\\
\midrule
\qquad\qquad\qquad\qquad\qquad\qquad Document as external knowledge (CMUDoG \cite{DBLP:conf/emnlp/ZhouPB18})\\
\midrule
\textcolor{blue}{\textbf{The Shape of Water}} is a 2017 American \textcolor{cyan}{\texttt{fantasy}} drama film directed by Guillermo del Toro ... It stars Sally Hawkins \\
... Set in Baltimore in 1962, the story follows a mute custodian at a high-security government laboratory who falls in \\
love with a captured humanoid amphibian creature. \textcolor{red}{\emph{Rating Rotten Tomatoes: 92 \% and average: 8.4/10}} ... Critical Res-\\
ponse: \textcolor{blue}{\textbf{one of del Toro's most stunningly successful works}}, a powerful vision of a creative master feeling totally\\
\midrule
Speaker 1: I thought \textcolor{blue}{\textbf{The Shape of Water was one of Del Toro's best works}}. What about you? \\
Speaker 2: Yes, his style really extended the story.\\
Speaker 1: I agree. He has a way with \textcolor{cyan}{\texttt{fantasy}} elements that really helped this story be truly beautiful. \\
Speaker 2: It has a \textcolor{red}{\emph{very high rating on rotten tomatoes}}, too. I don't always expect that with movies in this genre.\\
\bottomrule
  \end{tabular}
\end{table*}

%PersonalDialog
%In the PersonalDialog (PD) dataset which contains millions of dialogues (in single turn and multi turns) and various persona information for a large number of speakers. Specifically, this dataset comes from Weibo (a large social media in China). Each dialogue in it is composed of a Weibo post and its following replies. The personal profile of each speaker is collected and three persona attributes (i.e., Gender, Age, and Location) are approached in this study. 

Compared with structured knowledge such as the KG, unstructured text has the following advantages. \textbf{Firstly}, the unstructured text contains \textit{diversified information}, such as commonsense knowledge \cite{DBLP:journals/corr/abs-2010-03205}, stylized information \cite{DBLP:conf/emnlp/GaoZLGBGD19,DBLP:conf/acl/SuSZZHZNZ20}, syntactic structures \cite{DBLP:conf/naacl/DuB19}, and event logic \cite{DBLP:conf/ijcai/XuLWN0C20}. \textbf{Secondly}, unstructured text is \textit{ubiquitous}, \textit{continuously updated} and \textit{easy to obtain} due to the rapidly increased web-page content \cite{DBLP:journals/corr/abs-2004-14162}. These advantages indicate that unstructured text owns a greater potential to serve as external knowledge in DS than structured knowledge. Hence, incorporating unstructured text into dialogue attracted the attention of academia (e.g. the Alexa Prize challenge \cite{DBLP:journals/corr/abs-1801-03604}, Dialogue System Technology Challenges (DSTC) \cite{DBLP:journals/corr/abs-1901-03461,DBLP:journals/corr/abs-2101-09276}) and industry \cite{DBLP:journals/corr/abs-2004-13637,DBLP:conf/www/SzpektorCEFHKKO20,DBLP:conf/sigdial/KimEGHLH20,DBLP:journals/corr/abs-2009-03457}. Thanks to the progressing of the semantic representation learning technology such as pre-trained models \cite{DBLP:conf/naacl/PetersNIGCLZ18,DBLP:conf/naacl/DevlinCLT19,radford2018improving,radford2019language,DBLP:journals/corr/abs-1907-11692,DBLP:journals/corr/abs-1906-08237,DBLP:conf/nips/00040WWLWGZH19,DBLP:conf/acl/LewisLGGMLSZ20}, leveraging knowledge in unstructured text to build an informative and engaging dialogue agent have witnessed great improvement \cite{DBLP:conf/aaai/GhazvininejadBC18,DBLP:journals/corr/abs-1811-01241,DBLP:journals/corr/abs-2002-08909,DBLP:conf/www/SzpektorCEFHKKO20}. In this paper, we give a detailed investigation of the \textbf{Unstructured Text Enhanced Dialogue System (UTEDS)}.

%\subsection{Comparison with related literature reviews}%The tasks included DS, QA, question generation, summarization, machine translation, etc. The knowledge mode can be topic, keyword, knowledge base, knowledge graph, grounded text. %conditional-NLG %A Survey of Natural Language Generation Techniques with a Focus on

Although UTEDS is considered as a promising research direction, the earlier survey papers in DS research \cite{DBLP:journals/sigkdd/ChenLYT17,DBLP:journals/dad/SerbanLHCP18,DBLP:journals/jzusc/ShumHL18,DBLP:conf/ijcai/Yan18,DBLP:journals/corr/abs-1905-05709,DBLP:journals/corr/abs-1905-04071} did not cover this topic. Recent literature reviews in natural language generation (NLG) research noticed the emerging of UTEDS. \citet{DBLP:journals/corr/abs-1909-03409} focused on the conditional-NLG technology and introduced knowledge-enhanced text generation. \citet{DBLP:journals/corr/abs-1906-00500} focused on the NLG techniques in DS and introduced several models incorporating world knowledge into dialogue generation. These work only partially reviewed some related UTED models. Most recently, \citet{DBLP:journals/corr/abs-2010-04389} considered a variety of different knowledge-enhanced text generation tasks and answered two questions: how to acquire knowledge and how to incorporate different forms of knowledge to facilitate text generation. However, their investigation about free-form text grounded DS was also incomplete. In contrast, we focus on DS with unstructured text as external knowledge and give a detailed and systematic survey in this domain. We give definitions of the UTEDS related concepts, divide UTEDS into retrieval and generative models and summarize current models with a unified paradigm. We introduce the related datasets, specific architectures, methods, training objects, and evaluation metrics used in UTEDS. Then we summarize and analyze the experimental results of current models. In addition, we leverage the modules defined in this paper to point out the future research trends of UTEDS. 

As far as we know, we are the first to make a systematic review of the UTEDS. We believe that incorporating unstructured text information into dialogue generation is the future trend of the open-domain DS because a large amount of human knowledge is contained in these raw texts. The research of the UTEDS can assist machines in utilizing human knowledge stored on the internet and understanding natural language. 

The structure of the rest of the paper is as follows: 

\begin{itemize}
%\item In Section \textbf{Introduction}, we define and introduce UTEDS.
\item In Section \textbf{2}, we review the UTED datasets.
\item In Section \textbf{3}, we present the system components of Retrieval approaches. 
\item In Section \textbf{4}, we outline the system components of the Generative approaches. 
\item In Section \textbf{5}, we summarize the current evaluation metrics.
\item In Section \textbf{6}, we analyze the current models' performance in UTED.
\item In Section \textbf{7}, we put forward promising research directions.
\end{itemize}

%%%%%%%%%%%%%%%%%%%%%%%%%%%%%%%%%
\section{Datasets}

Recently, a number of UTED datasets based on different domains have been released \cite{DBLP:conf/acl/KielaWZDUS18,DBLP:conf/emnlp/ZhouPB18,DBLP:conf/emnlp/MogheABK18,DBLP:journals/corr/abs-1811-01241,DBLP:conf/acl/RashkinSLB19,DBLP:conf/acl/QinGBLGDCG19,DBLP:conf/interspeech/GopalakrishnanH19}. The background unstructured knowledge comes from multi-sources (e.g. Wikipedia, Newsreports) and the dialogues are either collected from crowd-sourcing platform (Amazon Mechanical Turk, AMT) or crawled from websites such as Reddit.com. 

Table \ref{diff_of_datasets} summarizes the characteristics of released UTED datasets from 6 aspects: the source of the external text, which side of the interlocutors can see the external text, source of conversation, the domain of dialogues, whether the knowledge texts in the dataset have been labeled, which side leads the conversation. We give a brief introduction of these datasets based on the classifications we defined in Tabel \ref{Classification}.

\begin{table}[h]
\footnotesize
\caption{Comparison of UTED datasets. * means the external knowledge of different interlocutors can be different. \# means the dataset could be used as UTED. "-" means the text is crawled independently from the dialogue. "Know." is short for Knowledge. "Philo\&Liter" means philosophy and literature. "Flu." means fluency. (\textbf{m.l.}) means the dataset is multi-lingual, while others are all English. "Fours." stands for Foursquare tips. "Gov." stands for Government.}
\label{diff_of_datasets}
\begin{tabular}{l|c|c|c|c|c|c}
\toprule
Dataset (Language) & Know. Source & See Text & Dialogue Source & Domain & Labeled & Lead by \\
\midrule
ARW \cite{DBLP:conf/coling/VougiouklisHS16}      & Wikipedia & - &Reddit Comments& Philo\&Liter & No & Both \\%sentence
GCD \cite{DBLP:conf/aaai/GhazvininejadBC18}     & Fours./Twitter & - &Twitter& Comments & No & User\\%
Persona-Chat \cite{DBLP:conf/acl/KielaWZDUS18}& AMT & Both* & ParlAI(AMT) &Persona& No&Both\\
X-Persona(\textbf{m.l.}) \cite{DBLP:journals/corr/abs-2003-07568}   & AMT & Both* & TranslationAPI &Persona & No &Both\\
M-Persona \cite{DBLP:conf/emnlp/MazareHRB18} & Reddit & Both &Reddit & Open & No & Both\\
%PD~ \cite{DBLP:journals/corr/abs-1901-09672}&WeiboPost& Both & Weibo& Open&No & Both\\~ means the knowledge is several words.
ED\# \cite{DBLP:conf/acl/RashkinSLB19}       & AMT & User & AMT & Open & No & User\\
CMUDoG \cite{DBLP:conf/emnlp/ZhouPB18}  &Wikipedia  & User/Both & AMT & Movie & No & Both \\
Holl-E \cite{DBLP:conf/emnlp/MogheABK18}  &Wikipedia & Both & AMT & Movie & Span \& Flu. & Both \\%fluency
WoW \cite{DBLP:journals/corr/abs-1811-01241} &Wikipedia &Both & ParlAI(AMT) & Open & Know. & User \\
CbR \cite{DBLP:conf/acl/QinGBLGDCG19}     &Wikipedia      & All & Reddit & Open & No & All\\
T-Chat \cite{DBLP:conf/interspeech/GopalakrishnanH19} & Multi Source & Both*& ParlAI(AMT) & Open & No & Both\\
LIGHT \cite{DBLP:conf/emnlp/UrbanekFKJHDRKS19}       &AMT & Both & ParlAI(AMT) & Game & No & Both \\
BST \cite{DBLP:conf/acl/SmithWSWB20}    & ParlAI(AMT) &  Both* & ParlAI(AMT) & Open & Dialogue Mode & Bot\\
KOMODIS \cite{DBLP:conf/lrec/GaletzkaES20}   &IMDB&Both*&AMT&Movie& Entities, etc. & User \\
Doc2Dial\# \cite{DBLP:conf/emnlp/FengWGPJL20}    & Gov. Website &  Bot & Doc2Dial \cite{DBLP:conf/aaai/FengFLL20} & Gov. service & Know., etc. & User\\
Interview \cite{DBLP:conf/emnlp/MajumderLNM20} & Public Radio & Bot & Public Radio & Open & Dialogue acts & Bot \\
Kialo \cite{DBLP:conf/emnlp/ScialomTSG20} & kialo.com & Both & kialo.com & Open & No & Both \\
PEC \cite{DBLP:conf/emnlp/ZhongZWLM20} & Reddit  & Both  & Reddit & Open & No & Both \\
%language quality, naturalness and attentiveness. “My partner is a native speaker of English”,“This felt like a natural chat about movies”,“My partner chatted like an attentive person would”. • Naturalness: The conversation felt natural. • Attentiveness: It felt like your chat partner was attentive to the conversation. • Consistency: The conversation was overall consistent. • Personality: The conversation fits well to the intended character described above. • Knowledgeability: Your partner replied correct to the asked questions.
\bottomrule
\end{tabular}
\end{table}

1) \textbf{Factual description:} \citet{DBLP:conf/coling/VougiouklisHS16} first proposed a dataset Aligning Reddit comments with Wikipedia sentences (ARW) and a task to generate context-sensitive and knowledge-driven dialogue responses. To provide the conversation model with relevant long-form text on the fly as a source of external knowledge, \citet{DBLP:conf/acl/QinGBLGDCG19} proposed CbR task where the dialogue utterance is accompanied with web-text. It is also used by the Sentence Generation Track of DSTC-7 \cite{DBLP:journals/corr/abs-1901-03461}. Topical-Chat \cite{DBLP:conf/interspeech/GopalakrishnanH19} dataset relies on multiple data sources, including Washington Post articles, Reddit fun facts, and Wikipedia articles about pre-selected entities, to enable interactions between two interlocutors with no explicit roles. The external knowledge provided to interlocutors could be the same or not, leading to more diverse conversations. Recently, \citet{DBLP:journals/corr/abs-2005-12529} presented an amended version of the Topical-Chat dataset with dialogue action plan annotations on multiple attributes (knowledge, topic, dialogue act). The Doc2Dial \cite{DBLP:conf/emnlp/FengWGPJL20} uses documents from government websites as grounded knowledge to imitate dialogues in government service. The dataset was collected and processed with an end-to-end framework \cite{DBLP:conf/aaai/FengFLL20}. Although it was proposed as a task-oriented dataset, it is also suitable for the UTED tasks.

2) \textbf{Fictional information:} The CMUDoG \cite{DBLP:conf/emnlp/ZhouPB18} is a dataset built with AMT where dialogue is generated based on the given background text (Wikipedia articles). The CMUDoG places no limits on the dialogues, except the two interlocutors respectively play an implicit recommender and recommended role. The Holl-E dataset \cite{DBLP:conf/emnlp/MogheABK18} was proposed to address the lack of external knowledge in the DS, which uses Reddit Comments, IMDB, and Wikipedia introductions as the background. There are two versions of the test set in the Holl-E: one with a single golden reference and the other with multiple golden references. The dialogues in the Holl-E allow one speaker to freely organize the language, while the second speaker needs to copy or modify the existing information (add words before or after a span to ensure smooth dialogue). It is worth noting that the background document has different length settings (mixed, oracle-reduced, oracle, full-reduced, full) in the dataset, please refer to the original paper for more detail.

\citet{DBLP:conf/acl/KielaWZDUS18} crowdsourced persona descriptions as external knowledge to construct a persona consistent dialogue (Persona-Chat). X-Persona dataset is proposed to create multilingual conversational benchmarks. In X-Persona, the training sets are automatically translated using translation APIs with several human-in-the-loop passes of mistake correction. In contrast, the validation and test sets are annotated by human experts to facilitate both automatic and human evaluations in multiple languages. In the Empathetic-Dialogue (ED) dataset, each conversation is grounded in a situation that is written by one participant following a given emotion label. There are 32 emotion labels. The person who wrote the situation description (Speaker) initiates a conversation to talk about it. The other conversation participant (Listener) becomes aware of the underlying situation through what the Speaker says and responds. The WoW dataset \cite{DBLP:journals/corr/abs-1811-01241} is constructed with ParlAI \cite{DBLP:conf/emnlp/MillerFBBFLPW17} to fill the absence of a supervised learning benchmark task on knowledgeable open dialogue with clear grounding. Each conversation happens between a wizard who has access to knowledge (paragraphs and sentences of articles) about a specific topic and an apprentice who is just eager to learn from the wizard about the topic. Learning in Interactive Games with Humans and Text (LIGHT) \cite{DBLP:conf/emnlp/UrbanekFKJHDRKS19} is a large-scale crowdsourced text adventure game, which is used as a research platform for studying grounded dialogue. Within that game world, the authors collected a large set of character-driven human-human interactions involving actions, emotes, and dialogues, with the aim of training models to engage humans in a similar fashion. In the BlendedSkillTalk (BST) \cite{DBLP:conf/acl/SmithWSWB20} dataset, the authors provided responses from poly-encoder \cite{DBLP:conf/iclr/HumeauSLW20} models that have been trained towards specific skills (ConvAI2, ED, and WoW) as reference to workers in the conversation.

3) \textbf{Subjective comments:} \citet{DBLP:conf/aaai/GhazvininejadBC18} leveraged Foursquare tips from a different person as external knowledge for Twitter conversation, namely Grounded Conversation Datasets (GCD). Besides plots and articles, datasets such as the DSTC-7 track-2, CMUDoG, and Holl-E also used movie reviews or Reddit comments and external knowledge. The subjective comments often accompany by factual descriptions. For example, in the Knowledgeable and Opinionated MOvie DIScussions (KOMODIS) dataset \cite{DBLP:conf/lrec/GaletzkaES20}, every dialogue is constrained by a unique set of facts as well as a suitable set of opinions about the entities in the facts. The participants rated their partner in terms of Naturalness/Attentiveness/Consistency/Personality/Knowledgeability as labels. The CMUDoG \cite{DBLP:conf/emnlp/ZhouPB18} and the Holl-E \cite{DBLP:conf/emnlp/MogheABK18} both include subjective comments. Interview \cite{DBLP:conf/emnlp/MajumderLNM20} dataset is collected from National Public Radio, the dialogues are news interviews based on given situational context. Kialo \cite{DBLP:conf/emnlp/ScialomTSG20} dataset is collected from www.kialo.com and uses stance-persona as external knowledge. M-persona \cite{DBLP:conf/emnlp/MazareHRB18} and PEC \cite{DBLP:conf/emnlp/ZhongZWLM20} construct persona descriptions from Reddit comments. 

\begin{table}[h]
\footnotesize
\caption{Statistics of UTED datasets. * means the statistic is calculated by us. Bold font indicates the maximum value in the column.}
\label{statistics_of_datasets}
\begin{tabular}{l|c|c|c|c|c}
\toprule
dataset & Dialogues & Turns/Dialogue & Words/Turn &Number.of.Text & Words/Text  \\
\midrule
ARW \cite{DBLP:conf/coling/VougiouklisHS16}      & 15,457* & 5.0* & \textbf{57.7}* & 75,171*(sentences) & 23.7* \\%sentence
GCD \cite{DBLP:conf/aaai/GhazvininejadBC18}      & 1,063,503 & 2.0 & 16.7& 43,271,508 (facts) & 17.6 \\%from HybridNCM_twitter_Foursquare
Persona-Chat \cite{DBLP:conf/acl/KielaWZDUS18} & 11,981 & 15.0 & 13.7* & 1,155 (persona) & 27.25* \\
X-Persona \cite{DBLP:journals/corr/abs-2003-07568}&123,484* &14.7* & 15.4* & 555,185*(persona)& 9.77*\\
%PD\cite{DBLP:journals/corr/abs-1901-09672}& 20.83M & 2.7 & 8.47M (Persona) & - & 9.4 \\means statistics of the training data.
M-Persona \cite{DBLP:conf/emnlp/MazareHRB18} &\textbf{700M}&-&-&\textbf{5M} (persona)&-\\
ED \cite{DBLP:conf/acl/RashkinSLB19}           & 24,850 & 4.3 & 15.2  & 24,850 (emotion \& situation) & 19.8\\
CMU\_DoG \cite{DBLP:conf/emnlp/ZhouPB18}   & 4,112 & 21.4 & 18.6 & 120 (documents) & 908 \\%The average number of turns are calculated as the number of utterances divided by the number of conversations for each of the datasets.
Holl-E \cite{DBLP:conf/emnlp/MogheABK18}   & 9,071 & 10.0 & 15.3 & 921 (documents) & 727.8 \\%one/turn
WoW \cite{DBLP:journals/corr/abs-1811-01241} & 22,311 & 9.1 & 17.2* & 1,356,509 (sentences) & 30.7 \\%one/turn%5.4M wikipedia articles. sentences knowledge: 60.8/turn. 30.7words/sentence. 17.2 doubtable
CbR \cite{DBLP:conf/acl/QinGBLGDCG19}      & 2.82M & \textbf{86.2} & 18.7 & 32.7k (documents) & \textbf{7,347.4} \\
T-Chat \cite{DBLP:conf/interspeech/GopalakrishnanH19} & 11,319 & 21.9& 19.7  & 3,064* (documents) & - \\%830*
LIGHT \cite{DBLP:conf/emnlp/UrbanekFKJHDRKS19}       & 10,777 & 13.0  & 18.3  & 10,777 (background) & - \\
BST \cite{DBLP:conf/acl/SmithWSWB20}        & 6,808 & 11.2* &  - & 6,808 (document) & - \\
KOMODIS \cite{DBLP:conf/lrec/GaletzkaES20}   & 7,519 & 13.8 & 14.4 & 13,818 (facts \& opinions)& -  \\
Doc2Dial \cite{DBLP:conf/emnlp/FengWGPJL20}    & 4,470 & 15.6* & 14 & 458 (documents) & 947 \\
Interview \cite{DBLP:conf/emnlp/MajumderLNM20} & 105,848 & 30.2 & - & 105,848 (situation) & -\\
Kialo \cite{DBLP:conf/emnlp/ScialomTSG20}& 241,882 & 2 & - & 18,255 (persona) & - \\% 241,882
PEC \cite{DBLP:conf/emnlp/ZhongZWLM20} & 355K& 2.3* & 25.5* & 250K (persona) & 10.9 \\
\bottomrule
\end{tabular}
\end{table}

Table \ref{statistics_of_datasets} illustrates the statistics of the above-mentioned UTED datasets with total dialogue numbers, average turns per dialogue, average word of each utterance, the total number of unstructured texts, average words of each unstructured text. "Turn" represents one utterance of a speaker, two turns make a conversation. We can observe that, besides Words/Turn, there are big gaps between datasets in the rest metrics. In order to make better use of these data, the existing UTED models \citep{DBLP:conf/ijcai/ZhaoTWX0Y19,DBLP:conf/acl/LiNMFLZ19,DBLP:conf/ksem/TangH19,DBLP:journals/corr/abs-1906-06685,DBLP:journals/corr/abs-1908-09528,DBLP:journals/corr/abs-1908-06449,DBLP:journals/corr/abs-1903-10245,DBLP:conf/naacl/AroraKR19,DBLP:conf/iclr/KimAK20,DBLP:conf/iclr/ZhaoWTX0020,DBLP:conf/aaai/Sun0XYX20} adopted different structures and methods. In the following two sections, we categorize the UTEDS into Retrieval models and Generative models and analyze the current approaches from a component perspective. Although these models conducted experiments on UTED datasets with given texts, they can be easily generalized to online models with a decent Information Retrieval (IR) module collecting related knowledge \cite{DBLP:conf/acl/ChenFWB17,DBLP:journals/corr/abs-2004-14162}.

%%%%%%%%%%%%%%%
\section{Retrieval models}
\begin{figure}[h]
\centering
\includegraphics[width=3.5in]{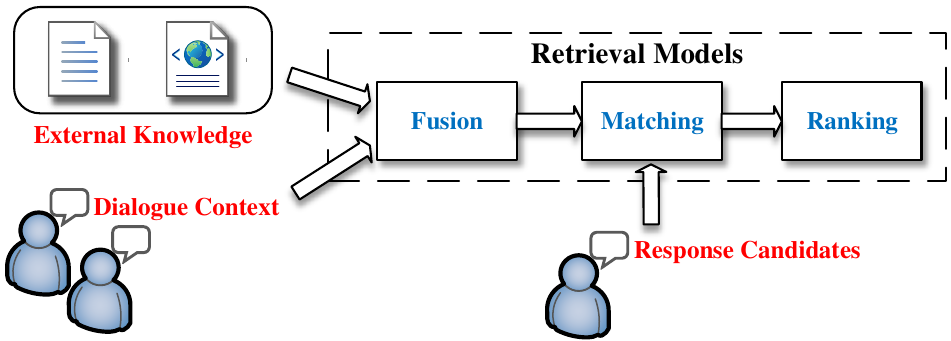}%\textwidth
\caption{The general system component of the Retrieval models in UTEDS.}
\label{UTED-R}
\end{figure}

Retrieval models produce a response by selecting from a candidate set. The advantage of the retrieval models is that it can give fluent responses, but the disadvantage is that the candidates are pre-defined, which causes the retrieval models hard to make full use of the dialogue context and external knowledge. Their structures are shown in Figure \ref{UTED-R}, the retrieval models are compose of three modules: \textbf{Fusion}, \textbf{Matching} and \textbf{Ranking}, 

Retrieval models take the external knowledge $\textbf{K}$ = ($K_1$, $K_2$, ..., $K_{|K|}$) with $|K|$ sentences, dialogue context $\textbf{C}$ = ($C_1$, $C_2$, ..., $C_{|C|}$) with $|C|$ utterances, and response candidates $\textbf{R}$ = ($R_1$, $R_2$, ..., $R_{|R|}$) with $|R|$ candidates as input, aim to select a best response $R_i$, ($i \in \{1, 2, ..., |R|\}$) by computing matching scores over the interaction results of \textbf{K}, \textbf{C}, and \textbf{R}. The Retrieval models first encode \textbf{K}, \textbf{C}, and \textbf{R} to get the hidden representations $h_K$, $h_C$, and $h_R$. The encoders\footnote{For example, HAKIM \cite{DBLP:conf/aaai/Sun0XYX20} used BIGRU + self-attention, FIRE \cite{DBLP:journals/corr/abs-2004-14550} used BILSTM to compute the sentence representations.} for each input can share parameters \cite{DBLP:journals/corr/abs-2004-14550} or not \cite{lowe2015incorporating,DBLP:conf/ijcai/ZhaoTWX0Y19,DBLP:conf/aaai/Sun0XYX20}. After obtaining the hidden representations, the \textbf{fusion} layer performs interactions among $h_K$ and $h_C$, then \textbf{matching} layer matches each candidates with the fusion results. Lastly, the \textbf{Ranking} layer ranks all the candidates with the matching results.
%DGMN uses Glove to retrain an embedding on task data. document and context are treated as separated sentences
%HAKIM uses BIGRU (Glove) and self-attention(compute sentence representation) as the encoder. it also uses separated sentences. 
%FIRE uses Glove+W2V+Character(CNN), BILSTM(for sentence representation) it also uses separated sentences.

\subsection{Fusion}
The Fusion layer fuses the information of $\textbf{K}$ and $\textbf{C}$ for the matching layer. Following \citet{DBLP:conf/cikm/GuoFAC16} who classified deep matching models into interaction-focused and representation-focused, we categorize the fusion methods into \textbf{interaction-based} and \textbf{representation-based}\footnote{Notably, the definition of ours is not exactly the same as \citet{DBLP:conf/cikm/GuoFAC16}, please refer to their paper for more details.}. 

\textbf{Interaction-based} method obtains multiple representation matrices from $h_K$ and $h_C$. The interaction can be different forms of attention mechanisms \cite{DBLP:journals/corr/BahdanauCB14,DBLP:conf/nips/VaswaniSPUJGKP17,DBLP:journals/corr/abs-2005-00743}. For example, \citet{DBLP:conf/ijcai/ZhaoTWX0Y19} conducted experiments on Persona-Chat and CMU\_DoG datasets. In their Document-grounded Matching Network (DGMN), document-aware context matrices $h^K_C$ and context-aware document matrices $h^C_K$ are interaction results of the encoding results $h_K$ and $h_C$. \citet{DBLP:journals/corr/abs-2004-14550} also employed the interaction-based fusion to get the knowledge-filtered context and context-filtered knowledge.

\textbf{Representation-based} method focuses on learning a better semantic representation such as a fusion vector for computing a simple similarity with response candidates. \citet{DBLP:conf/acl/KielaWZDUS18} proposed a Profile-Memory model learning to integrate weighted knowledge into context representation. The augmented context representation was used to match the response candidates. \citet{DBLP:conf/aaai/Sun0XYX20} adopted a History-Adaption Knowledge Incorporation Mechanism (HAKIM) which took the i-th turn history representation $h^{i}_C$ and knowledge representation $h^{i}_K$ as input and output the updated knowledge representation $h^{i+1}_K$ and knowledge-aware history representation $h^{i}_{C_K}$. For the (i+1)-th turn context, $h^{i+1}_C$ and $h^{i+1}_K$ were taken as input until all the context representations were updated. Then they employed a hierarchical encoder to encode all knowledge-aware context representations into a history vector.

\subsection{Matching}
The Matching layer pairs the fusion results with each candidate and extracts matching information from the pairs. It can be categorized into \textbf{shallow matching} and \textbf{deep matching}. 

\textbf{Shallow matching} method performs a single interaction after fusion. For example, \citet{lowe2015incorporating} proposed a Knowledge Enhanced (KE) model with the Ubuntu dialogue Corpus as dialogue and the Ubuntu Manpages as external knowledge. They extended the dual-encoder model \cite{DBLP:conf/sigdial/LowePSP15} by separately encoding the \textbf{K}, \textbf{C}, and $\textbf{R}_i$ into hidden layer representations $h_K$, $h_C$, and $h_{R_i}$. Then they used $\sigma$ (${h_k}^T  M_k  h_{R_i} $+ ${h_c}^T  M_c  h_{R_i}$) to compute the matching score of $R_i$. Where $M_k$ and $M_c$ were learnable parameters. \citet{DBLP:conf/ijcai/ZhaoTWX0Y19} also employed the shallow matching except $h_{R_i}$ was interacted with 3 items: context $h_C$, knowledge-aware context $h^K_C$ and context-aware knowledge $h^C_K$. The 3 interaction matrices were aggregated into a vector with Multilayer perceptron (MLP) to calculate the final matching score. \citet{DBLP:conf/sigir/YangQQGZCHC18} proposed a Deep Matching Networks (DMN) for information-seeking conversation. The external knowledge was QA pairs extracted from a Knowledge Base using response candidates. The DMN calculated the matching information between each dialogue turn and the candidate response and used a Gated Recurrent Unit (GRU) \citet{DBLP:conf/emnlp/ChoMGBBSB14} or MLP to aggregate all the sequence matching information into a vector. The vector was finally utilized to compute the score of this candidate.

\textbf{Deep matching} method performs a iteratively referring after fusion. For example, \citet{DBLP:journals/corr/abs-2004-14550} proposed a Filtering before Iteratively REferring (FIRE) Model. They first employed the interaction-based fusion to get the knowledge-filtered context $h^K_C$ and context-filtered knowledge $h^C_K$. On one side, they interacted $h_R$ with $h^K_C$ to get first layer refering results $h^{K_C,1}_R$ and $h^{R,1}_{K_C}$. Then $h^{K_C,1}_R$ and $h^{R,1}_{K_C}$ were interacted to get the second layer refering results $h^{K_C,2}_R$ and $h^{R,2}_{K_C}$. On the other side, $h^C_K$ was interacted with $h_R$ in the same way to obtain first layer refering results $h^{C_K,1}_{R}$ and $h^{R,1}_{C_K}$. $h^{K_C,1}_R$, $h^{R,1}_{K_C}$, $h^{C_K,1}_{R}$ and $h^{R,1}_{C_K}$ were used to calculate the first layer matching score $g^1$. In their experiments, the iteration layer was set to 3 and the final score was the average of 3 matching scores.

\subsection{Ranking}
There are three commonly used training methods for ranking: Point-wise, Pair-wise, and List-wise. Only the first two are employed by UTED models at present. \textbf{Point-wise} calculates the scores of all candidate responses independently \cite{DBLP:conf/acl/KielaWZDUS18,DBLP:journals/corr/abs-2004-14550}. It usually employs Cross-Entropy Loss as an objective function. One disadvantage is that the relative relationship between different candidates is not considered. \textbf{Pair-wise} considers the relative relationship between different candidates. Models using Pair-wise ranking normally adopt hinge loss function \cite{DBLP:conf/sigir/YangQQGZCHC18,DBLP:conf/acl/KielaWZDUS18} to distinguish a positive candidate with a negative one and learn to score higher for the ground-truth response. One disadvantage is that the distribution of all candidates is not modeled. In contrast, List-wise directly models the distribution of all response candidates and employs Kullback-Leibler (KL) Divergence to optimize the gap between the generated distribution and the ground truth one. In Table \ref{Retrieval}, we summarize the current Retrieval-based models with the Fusion/Matching/Ranking methods.

\begin{table}[h]
\footnotesize
\caption{Module comparison between different Retrieval models. "-" means no fusion operation. "R-b" stands for Representation-based. "I-b" means Interaction-based. "Point/Pair" are short for Point/Pair-wise, respectively.}
\label{Retrieval}
\begin{tabular}{l|ccc|l|ccc}
\toprule
Models  & Fusion & Matching & Ranking  & Task &  Fusion & Matching & Ranking \\
\midrule
TF-IDF baseline \cite{DBLP:conf/acl/KielaWZDUS18}     & - & Shallow & Point &TMN \cite{DBLP:journals/corr/abs-1811-01241}  & - & Shallow & Point\\%tf-idf weighted cosine similarity (cosine is also a interaction)
Starspace \cite{DBLP:conf/acl/KielaWZDUS18}       & - & Shallow &  Pair &  Transformer \cite{DBLP:conf/emnlp/MazareHRB18}  & R-b & Shallow & Point  \\%K concat to C
KV-Profile Memory \cite{DBLP:conf/acl/KielaWZDUS18}      & R-b  & Shallow  & Point &DGMN \cite{DBLP:conf/ijcai/ZhaoTWX0Y19}  & I-b & Shallow & Point  \\%K concat to C
KE \cite{lowe2015incorporating}     & R-b & Shallow & Point &HAKIM \cite{DBLP:conf/aaai/Sun0XYX20}     & R-b & Shallow & Point  \\
Retrieval-TMN \cite{DBLP:journals/corr/abs-1811-01241}     & R-b & Shallow & Point  & FIRE \cite{DBLP:journals/corr/abs-2004-14550}  & I-b & Deep & Point \\
%DGMN uses Glove to retrain an embedding on task data. document and context are treated as separated sentences
%HAKIM \cite{DBLP:conf/aaai/Sun0XYX20} uses BIGRU (Glove) and self-attention(compute sentence representation) as encoder. it also uses seperated sentences. 
%FIRE \cite{DBLP:journals/corr/abs-2004-14550} uses Glove+W2V+Character(CNN), BILSTM(for sentence representation) it also uses seperated sentences. 
%DMN-PRF \cite{DBLP:conf/sigir/YangQQGZCHC18}DMN-KD \cite{DBLP:conf/sigir/YangQQGZCHC18}
\bottomrule
\end{tabular}
\end{table}

%%%%%%%%%%%%%%%%%%%%%%%%%%%%%%%%%%%%%%

%%%%%%%%%%%%%%%%%%%%%%%%%%%%%%%
%%%%%%%%%%%%%%%%%%%%%%%%%%%%%%%
\section{Generative Models}
%In this section, we classify the generative models into 3 modules, analyze their motivations and connections between each other. %Focus on its different characteristics from traditional data-driven dialogue systems

\begin{figure}[h]
\centering
\includegraphics[width=4in]{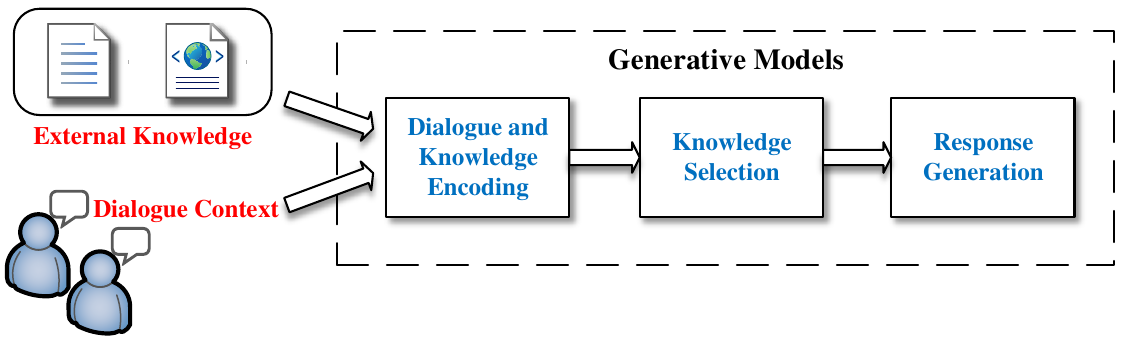}
\caption{The general system component of the Generative models in UTEDS.}
\label{UTED-G}
\end{figure}

Generative models generate novel sentences that are more natural and variable by conditioning on the dialogue history and external knowledge. Their structures are shown in Figure \ref{UTED-G}, the generative models consist of \textbf{Dialogue and Knowledge Encoding}, \textbf{Knowledge Selection}, and \textbf{Response Generation}. 

The UTED generation task can be defined as follows: given unstructured text $\textbf{K}$ = ($K_1$, $K_2$, ..., $K_{|K|}$) with $|K|$ sentences and dialogue context $\textbf{C}$ = ($C_1$, $C_2$, ..., $C_{|C|}$) with $|C|$ utterances, the task is to generate a $r$ tokens response \textbf{R} = ($R^1$, $R^2$, ..., $R^r$) with probability: $P(\textbf{R} | \textbf{C}, \textbf{K} ; \Theta ) = \prod_{i=1}^r P(R^i | \textbf{C}, \textbf{K}, \textbf{R}^{<i}; \Theta)$, where $\textbf{R}^{<i} = (R^1, R^2, ..., R^{i-1})$ and $\Theta$ is the model's parameters. The \textbf{Dialogue and Knowledge Encoding} module encodes \textbf{K}, \textbf{C} to get hidden representations $h_K$ and $h_C$. After obtaining the hidden representations, the \textbf{Knowledge Selection} layer selects useful knowledge from the memory constructed by $h_K$ and $h_C$, then \textbf{Response Generation} layer produces the response word by word with selected knowledge. There is also work trying to combine the advantages of retrieval and generative methods \cite{DBLP:conf/emnlp/WestonDM18}, where the authors first retrieved a response from candidates and then used the generation model to transfer the retrieval sentence into a more natural one. %The ground-truth response \textbf{R} can be used as the posterior information in the Teacher-Student training paradigm.%there are three kinds of dialogue retrieval, one is WestonDM18 who use external knowledge helping select dialogue, then use the dialogue for generation. another treats other dialogues in other corpus as external knowledge, use candidates to retrieve more related dialogue to help decide which candidate are better, like DMN-KD. the third treats dialogue in the same task corpus as assist to filter the real external KG like "Improving Knowledge-Aware Dialogue Response Generation by Using HumanWritten Prototype Dialogues".

%Each $\textbf{D}_i$ comprises $T_{D_i}$ tokens, $\textbf{D}_i$ = ($D_i^{1}$, $D_i^{2}$, ..., $D_i^{T_{D_i}}$), each utterance $\textbf{C}_i$ comprises $T_{C_i}$ tokens, $\textbf{C}_i$ = ($C_i^{1}$, $C_i^{2}$, ..., $C_i^{T_{C_i}}$). 

%%%%%%%%%%%%%%%%%%%%%%% Knowledge and Dialogue Encoding
%%%%%%%%%%%%%%%%%%%%%%%
\subsection{Dialogue and Knowledge Encoding}
Dialogue and knowledge encoding aims to learn good representations from raw text to capture important information for subsequent modules. Researchers utilized different encoding strategies to achieve this purpose. We categorize the current encoding methods into 4 classes: 1) without pre-training, 2) with pre-trained word embedding, 3) with contextualized word embedding, and 4) with explicit knowledge expansion.

\subsubsection{Without Pre-training}
Some models used randomly initialized word embeddings and learned them from scratch during training. For example, on Holl-E task, \citet{DBLP:journals/corr/abs-1906-06685} used GRU \cite{DBLP:journals/corr/ChungGCB14} and \citet{DBLP:journals/corr/abs-1908-06449} employed LSTM \cite{DBLP:conf/nips/SutskeverVL14} as encoder; on CMU\_DoG, \citet{DBLP:conf/acl/LiNMFLZ19} used an incremental Transformer \cite{DBLP:conf/nips/VaswaniSPUJGKP17} encoder.

\subsubsection{With Pre-trained Word Embedding}
The traditional word embeddings (Word2Vec \cite{DBLP:journals/corr/abs-1301-3781}, GloVe \cite{DBLP:conf/emnlp/PenningtonSM14}, etc.) were obtained by pre-training on large corpus. These embeddings contained semantic information and could help the downstream NLP tasks. In UTED, they were utilized to initialize the dialogue context and the knowledge, and then input into contextualized encoder \cite{DBLP:journals/corr/abs-1908-09528,DBLP:conf/acl/KielaWZDUS18,DBLP:conf/ijcai/ZhaoTWX0Y19}. Some work chose to train new word embeddings with these methods. For example, to learn the representations of internet slangs and spoken English in target corpus, \citet{DBLP:journals/corr/abs-1903-09813} trained a 100 dimension word embeddings via GLoVe from conversations and facts. However, the traditional word embeddings are fixed and context-independent, they could not resolve the out-of-vocabulary (OOV) problem and the ambiguity of words in different contexts. So the contextualized word embedding was introduced. %\cite{DBLP:conf/aaai/GhazvininejadBC18,DBLP:conf/emnlp/ZhouPB18,DBLP:conf/emnlp/MogheABK18,}

%Pre-training methods has long been used in UTED \cite{lowe2015incorporating}, \citet{DBLP:conf/coling/VougiouklisHS16} pre-trained a CNN-based sentence classification model \cite{DBLP:conf/emnlp/Kim14} on a subset of the Wikipedia-sentences dataset to select a sentence based on the topic-keyword. \citet{DBLP:conf/ijcai/LianXWPW19} minimized the BOW loss for pre-training the knowledge manager in PostKS.
%Encoding method such as Byte Pair Encoder (BPE for oov) \cite{DBLP:journals/corr/abs-2004-12744,DBLP:conf/interspeech/GopalakrishnanH19} and Positional Embedding \cite{DBLP:conf/naacl/DevlinCLT19} (PE for long dependency) can help PTM learn more semantic information.
% Even with a smaller data size, pre-training on corpus which shares a similar distribution with the target domain can also benefit the target task \cite{DBLP:conf/acl/GururanganMSLBD20}.

\subsubsection{With Contextualized Word Embedding}
To obtain contextualized word embedding, some Pre-Trained Models (PTMs) (ELMo \cite{DBLP:conf/naacl/PetersNIGCLZ18}, GPT-1 \cite{radford2018improving}, BERT \cite{DBLP:conf/naacl/DevlinCLT19}, etc.) were introduced. These PTMs were first trained on a large corpus, then fine-tuned on specific tasks. The contextualized embedding has been proven to be better for the downstream NLP tasks \cite{DBLP:conf/sigir/MengRCSRTR20,DBLP:journals/corr/abs-2003-08271} than traditional word embedding.

\citet{DBLP:journals/corr/abs-2005-14315} used ELMo and BERT to obtain pre-trained words embedding with task corpus. \citet{DBLP:conf/acl/GolovanovKNTTW19} investigated the different architecture of UTEDS and proposed a Multi-Input (dialogue history, facts, and previously decoded tokens) model where a GPT-1 was duplicated to form an encoder-decoder structure. \citet{golovanov2020lost} used the Transformer-enhanced GPT-1 \cite{radford2018improving}, and fine-tuned it on Persona-Chat dataset. Pre-trained GPT-1 was also utilized by \citet{DBLP:journals/corr/abs-2004-14614} as the knowledge base and performed knowledge selection in it. \citet{DBLP:journals/corr/abs-2008-12918} proposed zero-resource knowledge-grounded conversation (ZRKGC) model with a pre-trained UNILM \cite{DBLP:conf/nips/00040WWLWGZH19}. Adapter-Bot \cite{DBLP:journals/corr/abs-2008-12579} used DialoGPT \cite{DBLP:conf/acl/ZhangSGCBGGLD20} as a shared encoding backbone for several independent downstream adapters. Each adapter learned to converse with certain skill. %A Bert \cite{DBLP:conf/naacl/DevlinCLT19} was fine-tuned as Dialogue Manager to automatically switch among all adapters.
%\citet{DBLP:conf/emnlp/ZhaoWXTZY20} used BERT as encoder and GPT as decoder

Some researchers proposed their own PTMs with Transformer structure. \citet{DBLP:journals/corr/abs-1811-01241} introduced Transformer Memory Network (TMN), a Transformer framework pre-trained with Reddit \cite{DBLP:conf/emnlp/MazareHRB18} and SQuAD \cite{DBLP:conf/emnlp/RajpurkarZLL16} then fine-tuned on WoW. \citet{DBLP:journals/corr/abs-2004-12744} introduced a Transformer Seq-to-Seq model pre-trained on Reddit comments. 

%They proposed a Decoupling method first de-coupling some knowledge-related information from the input and then re-coupling the information back to the input. 

\subsubsection{With Explicit Knowledge Expansion}
The pre-trained word embedding has been proven to contain semantic knowledge (grammatical \cite{DBLP:conf/acl/TenneyDP19,DBLP:journals/corr/abs-2004-14448}, syntactical \cite{DBLP:conf/iclr/TenneyXCWPMKDBD19}, etc.) and world knowledge \cite{DBLP:conf/emnlp/PetroniRRLBWM19}, these knowledge can be considered as implicit. However, to obtain more explicitly semantic features for downstream tasks, researchers have adopted many \textbf{explicit knowledge expansion} methods, such as named-entity recognition (NER) \cite{DBLP:conf/acl/QinGBLGDCG19}, part-of-speech (POS) \cite{DBLP:journals/corr/abs-2005-06128}, syntactic tree \cite{DBLP:conf/naacl/DuB19} and hand-rule features\footnote{Matching features based on the original word, lower case, lemma, sequence feature, etc.}.

\citet{DBLP:conf/www/ZhaoWHYCW20} defined label knowledge (conversation topics, structured knowledge facts in KGs, etc.) as additional knowledge that enhanced language representations in different aspects. They used a label knowledge detection module to get the additional label representation, then divided the type of knowledge into two specific categories (labeled and unstructured) during fine-tuning and handled them in different ways. \citet{DBLP:journals/corr/abs-2005-00613} proposed controllable grounded response generation (CGRG). First, they used rule-based extraction to retrieve the co-occurring keywords/phrases in the dialogue context and the external knowledge, then only used the knowledge sentences containing these keywords/phrases as external knowledge. They used these co-occurrence relationships to simplify the attention operations in the GPT-2 \cite{radford2019language}, thereby controlling effective information injection and generation. PEE \cite{DBLP:conf/ecai/XuLYRRC020} adopted a VAE-based topic model to conduct external persona information mining. \citet{DBLP:journals/corr/abs-2005-14315} used GCN to learn structured information. They tested 3 different graph-based structures (dependency, entity co-reference, and entity co-occurrence graphs). \citet{DBLP:journals/corr/abs-2010-03205} first trained a Commonsense Transformers Framework \cite{DBLP:conf/acl/BosselutRSMCC19}, then generated expansions sentence for each external knowledge facts. 

%need Example
%Other researchers \cite{DBLP:conf/acl/LinJHWC20,DBLP:journals/corr/abs-2004-12744,DBLP:journals/corr/abs-2008-12579} provided universal encoding method to seamlessly leverage diverse knowledge sources such as multi-modal data. %However, their method focused on utilizing different forms of knowledge, not knowledge expansion.

%%%%%%%%%%%%%%%%%%%%%%%%%%%%%%
\subsection{Knowledge Selection (KS)}%where is the ZRKGC and DRD and Meta-Learning
As the core component of Unstructured Text Enhanced Dialogue System, \textbf{knowledge selection} module aims to extract semantically and logically related information from the encoded text representation based on given dialogue history. Most UTED systems followed the \textit{Memory Network Framework} \cite{DBLP:journals/corr/WestonCB14,DBLP:conf/emnlp/MillerFDKBW16} with an attention mechanism \cite{DBLP:journals/corr/BahdanauCB14,DBLP:conf/nips/VaswaniSPUJGKP17} to dynamically read appropriate knowledge from the constructed memory. In terms of the existence of a sampling mechanism that explicitly selects the most relevant text fragments from given background knowledge, we divide knowledge selection into \textbf{implicit (soft) selection} and \textbf{explicit (hard) selection}.

\subsubsection{Implicit Selection}%definition (into a memory), the problem is linking context with the knowledge 
Early UTED system with \textbf{implicit knowledge selection} \cite{DBLP:conf/aaai/GhazvininejadBC18,DBLP:conf/emnlp/ZhouPB18,DBLP:conf/acl/KielaWZDUS18,noauthororeditor,DBLP:conf/emnlp/MogheABK18} usually employed the attentional Seq-to-Seq memory network which encoded the context and unstructured text respectively into a vector or a sequence of vectors as model memory and used the decoder hidden state as a query to attentively read the memory (or concatenate if the encoder output is a single vector) \cite{DBLP:journals/corr/abs-2005-03174}. However, this naive structure depended solely on the attention mechanism to conduct knowledge selection, which was too simple to efficiently link the context with related knowledge and extract salient information.

%matching between context and knowledge
To address this problem, some researchers employed some matching operations between context and knowledge before constructing the memory, we call this \textbf{early interaction}. Many of them borrowed idea from the Machine Reading Comprehension (MRC) task, such as \citet{DBLP:journals/corr/abs-1908-06449} from match-lstm \cite{DBLP:conf/iclr/Wang017a}, \citet{DBLP:conf/acl/QinGBLGDCG19,DBLP:journals/corr/abs-2005-06128} from SAN \cite{DBLP:conf/acl/GaoDLS18} and \citet{DBLP:conf/naacl/AroraKR19,DBLP:journals/corr/abs-1906-06685} from BiDAF \cite{DBLP:journals/corr/SeoKFH16}. Instead of predicting a span, they took advantage of the matching techniques in the MRC model, such as cross attention and matching matrix, to generate a document-length memory. They posited that early interaction between context and background knowledge could enable the model to better utilize the context information to pre-select relevant knowledge and integrate the most salient ones together. Following a similar idea, \citet{DBLP:conf/acl/LiNMFLZ19} proposed Incremental Transformer which employed a Transformer structure \cite{DBLP:conf/nips/VaswaniSPUJGKP17} with stacked self-attention and cross-attention layers to incrementally encode context and document into a compound memory. Based on ITDD, \citet{DBLP:conf/emnlp/MaZS020} considered the relationships between current turn and history turns, they designed a Compare Aggregate Transformer (CAT) to reduce the noise introduced by history turn. \citet{DBLP:journals/corr/abs-1908-09528} further claimed that a single token lacks a global perspective, hence they utilized the matching matrix and implemented an "m-size unfold\&sum" operation to select continuous spans from the background and form a topic transition vector which is used to direct knowledge selection in the decoding process.

%use different attention
Besides early interaction, different attention mechanism was also investigated. \citet{DBLP:conf/cikm/ZhengZ19} proposed an Enhanced Transformer Decoder (TED) that attentively read from context and knowledge representations. \citet{DBLP:conf/ecai/XuLYRRC020} proposed a multi-hop memory retrieval mechanism which stacked several attention layers together, the output of the former layer was used as the query for subsequent. \citet{DBLP:journals/corr/abs-1911-09728} investigated three different approaches (Concatenate, Alternate and Interleave) to combining context and knowledge encodings into a Transformer type decoder. They showed that the Interleave mechanism consistently outperformed concatenate and alternate with extensive experiments. Notably, the attention mechanism used in \citet{DBLP:conf/aaai/ZhengZHM20} employed the Alternate way.

\subsubsection{Explicit Selection}%definition
Attention mechanisms have been found to operate poorly over long sequences, as the mechanism was blurry and difficult to make fine-grained decisions \cite{DBLP:conf/acl/LewisDF18}. The significant model performance decline with longer sequences observed in \citet{DBLP:conf/emnlp/ZhouPB18} also verified this proposition. As a consequence, some scoring and sampling mechanism was proposed to select fragments (usually a sentence) from original background text and subsequently fed the corresponding encodings into the decoder in a similar way as the implicit selection. We define this process as the \textbf{explicit selection} mechanism. Another difference between Implicit and Explicit is that models with the latter can measure the accuracy of KS.

%scoring %\citet{DBLP:journals/corr/abs-2004-12744}
Similar to the attention mechanism, scoring aims to match dialogue context with each pre-segmented text piece respectively and generate a preference distribution over them. \citet{DBLP:journals/corr/abs-1811-01241} and \citet{DBLP:conf/ijcai/LianXWPW19} respectively encoded knowledge sentence and utterance sentences into a sentence embedding vector and applied dot product attention \cite{DBLP:conf/emnlp/LuongPM15} to derive the preference distribution. Other simple scoring methods included but were not limited to MLP attention \cite{DBLP:journals/corr/BahdanauCB14} in \citet{DBLP:conf/acl/BaoHWLW19}, Term Frequency Inverse Document Frequency (TF-IDF) similarity in \citet{DBLP:conf/interspeech/GopalakrishnanH19} and K-Nearest-Neighbors in \citet{DBLP:journals/corr/abs-2004-12744}. \citet{DBLP:journals/access/AhnLP20} classified the above methods as Reduce-Match strategy since they firstly aggregated the utterances into a single vector and then matched it with a knowledge vector. They argued that the coarse information condense mechanism made fine-grained matching difficult, and they further proposed the Match-Reduce strategy which firstly implemented fine-grained token-level matching operations to obtain matching features (like matching matrices) and then utilized convolution and pooling operations to aggregate these features into a scalar score. Instead of computing one single preference distribution, \citet{DBLP:conf/emnlp/ZhaoWXTZY20} regarded knowledge selection as a sequence prediction process that used LSTM to sequentially select knowledge sentences.

%supplementary
Besides dialogue context, some researchers attempted to utilize other \textbf{supplementary} information to facilitate explicit knowledge selection. \citet{DBLP:conf/iclr/KimAK20} and \citet{DBLP:conf/sigir/MengRCSRTR20} posited that tracking the knowledge usage in dialogue history would enable the model to capture the topic flow between multi-turns and reduce knowledge repetition. \citet{DBLP:conf/iclr/KimAK20} employed a separate GRU to encode formerly selected knowledge and concatenated the output hidden state with context embedding to conduct scoring. Similarly, \citet{DBLP:conf/sigir/MengRCSRTR20} explicitly modeled knowledge tracking distribution and used the knowledge sampled from tracking distribution to facilitate the derivation of knowledge shifting distribution, which was exactly the preference distribution over sentences. \citet{DBLP:conf/emnlp/ZhengCJH20}
argued that the difference between the knowledge sentence selected at different dialogue turns usually provides potential clues to KS. They compared knowledge candidates with the previously selected sentence and used the comparison results as guidance for the final selection. In addition, \citet{DBLP:journals/corr/abs-2010-03205} used Commonsense Transformers Framework \cite{DBLP:conf/acl/BosselutRSMCC19} to generate expansions for each persona sentence along the nine relation type that ATOMIC \cite{DBLP:conf/aaai/SapBABLRRSC19} provided. They posited that different utterances had a different preference over the nine relation type, hence they encoded the relation type as additional information to assist scoring. \citet{DBLP:journals/corr/abs-1903-10245} even abandoned the scoring mechanism, instead, they expanded the background text with an augmented graph and aligned each knowledge sentence with a graph vertex. Reinforcement Learning was used to train the reasoning policy over the augmented graph and the final absorbed state (vertex) with its corresponding sentence was selected as the knowledge fragment. 

%training process  % Jaccard similarity \cite{DBLP:journals/corr/abs-1908-09528}? \citet{DBLP:journals/corr/abs-2005-06128} ?
Compared with implicit selection, the training process of the explicit selection model deserves more attention since the sampling operation over categorical distribution is non-differentiable which causes that the back-propagation with simple Negative Log-Likelihood (NLL) generation loss can not update the parameters of preference distribution. When the training dataset contains the ground truth knowledge label, an auxiliary knowledge loss (\textit{i.e.} the cross-entropy loss between the predicted preference distribution and the true knowledge distribution) can be applied to train the scoring module \cite{DBLP:journals/corr/abs-1811-01241,DBLP:conf/iclr/KimAK20,DBLP:conf/sigir/MengRCSRTR20}. However, if golden knowledge is not provided, other solutions are required. The most naive one is to construct a pseudo-knowledge label with some keyword matching techniques like TF-IDF cosine similarity \cite{DBLP:journals/access/AhnLP20}, BM-25 \cite{DBLP:conf/emnlp/ZhengMZ20}, unigram F1 \cite{DBLP:conf/emnlp/ZhaoWXTZY20} and Jaccard similarity \cite{DBLP:journals/corr/abs-1908-09528}. Moreover, \citet{DBLP:conf/acl/BaoHWLW19} designed two reward functions, informativeness reward and coherence reward, on generated sentence and used policy gradient to train the explicit knowledge selection module with encoder and decoder parameters pre-trained and fixed. Notably, Gumbel-Softmax \cite{DBLP:conf/iclr/JangGP17} is a common technique for training categorical variables. But \citet{DBLP:conf/ijcai/LianXWPW19} claimed that most of the existing knowledge selection scoring method which based on semantic similarity was problematic since in a dialogue setting the knowledge might not be semantically related to given utterance but logically. Simple semantic similarity matching would produce a biased prior distribution over the predicted preference distribution which made efficient training difficult without a true knowledge label. To remedy this problem, researchers introduced a teacher-student training mechanism that defined a response-anticipated scoring module and generated a posterior preference distribution which was easier trained with Gumbel-softmax because the golden response was exposed to the model. The posterior preference distribution acted like a soft label that was used to train the original scoring module via a KL divergence loss during training. Similar idea was applied by \citet{DBLP:journals/corr/abs-2010-03205} and \citet{DBLP:journals/corr/abs-2005-06128} as well.

%%%%%%%%%%%%%%%%%%%%
%%%%%%%%%%%%%%%%%%%%  Response Generation
%%%%%%%%%%%%%%%%%%%%
\subsection{Response Generation}%add training is odd too. 
The Response Generation module takes the results of previous modules (Knowledge and Dialogue Encoding and Knowledge Selection) and generates the final dialogue response. In this section, we summarize the Architectures, Methods, and Loss Functions used in generative models. The Losses are calculated with the decoding results and are closely related to the Architectures and Methods. Therefore, we introduce them together in this section for the convenience of reading. %We first summarize the Architectures and Methods used in UTED, then introduce the Copy Mechanism and Loss Function according to the difference between UTEDS and the traditional data-driven DS \cite{DBLP:journals/dad/SerbanLHCP18}. Although they require global considerations in both encoding and decoding parts, but they are more closely related to the decoding and training process than encoding and KS and are usually used in conjunction with each other. and is convenient for readers to read

\subsubsection{Architectures and Methods}
The Architectures and Methods utilized in UTED include: %but not limit to

\textbf{1) Sequence-to-Sequence (Seq-to-Seq)} architecture \cite{DBLP:conf/nips/SutskeverVL14,DBLP:conf/emnlp/ChoMGBBSB14} takes a sequence of words and generates a sequence word by word. Most of the methods used in UTED are based on Seq-to-Seq architecture. RNN-based \cite{DBLP:journals/corr/abs-1908-06449,DBLP:journals/corr/abs-1908-09528} and Transformer-based \cite{DBLP:journals/corr/abs-2004-12744,DBLP:conf/cikm/ZhengZ19} Seq-to-Seq architectures were widely used in UTED. \citet{DBLP:conf/acl/SongWZLL20} introduced a Generate, Delete and Rewrite framework for persona consistent dialogue generation. They adopted Seq-to-Seq Transformer architecture and integrated a matching model to delete the inconsistent words.
 
%Knowledge are concatenated to the dialogue context \cite{DBLP:journals/corr/abs-1811-01241,DBLP:conf/acl/LiRKWBCW20} or fused into joint representations \cite{DBLP:conf/acl/LiNMFLZ19,DBLP:journals/corr/abs-1908-09528}. 
%more details or used attention to integrated into context then into decoder

\textbf{2) Reinforcement Learning (RL)} agents make decisions and take actions serially through interaction with the environment, and obtain reward to train action strategy. \citet{DBLP:conf/emnlp/LiMRJGG16} first employed RL for dialogue generation, where a simple Seq-to-Seq model was used as the generation model. Three rewards were defined and combined together to boost diverse response generation. One advantage of RL is the method can leverage non-differentiable rewards. However, the model performance is sensitive to the design of the rewards and is hard to converge.

In UTED, \citet{DBLP:conf/acl/BaoHWLW19} introduced the Generation-Evaluation (GE) framework for UTED with the objective of letting both participants know more about each other. For the sake of rational knowledge utilization and coherent conversation flow, a dialogue strategy that controlled knowledge selection was instantiated and continuously adapted via reinforcement learning. Under the deployed knowledge selection strategy, two dialogue robots introduced themselves to each other based on their background information and responded appropriately to the utterances of both parties. The dialogues they generated and the corresponding background information were used for the strategy evaluation module to measure informativeness and coherence. These assessments were integrated into a compound reward, which served as a reinforcement signal to guide the continuous evolution of dialogue strategies. \citet{DBLP:conf/ijcai/XuLWN0C20} presented a novel event graph grounded RL framework (EGRL). The EGRL first constructed an event graph where vertices represented events (most simply verb phrases) and edges indicated relations (temporal order, causal relation, etc.) between the events, then used RL based multi-policy method to conduct high-level content planning. \citet{DBLP:conf/acl/LiuCCLCZZ20} proposed a Transmitter-Receiver based framework to explicitly model the understanding between interlocutors, where Transmitter (GPT) was responsible for dialogue generation, and Receiver (BERT) was responsible for personality understanding. The latter was similar to a discriminator, which determined whether the dialogue generated by the former fulfilled the requirements. A weighted sum of Language Style, Discourse Coherence, and Mutual Persona Perception was calculated as a reward for RL.

%Deep reinforcement learning for chatbots using clustered actions and human-likeness rewards

\textbf{3) Conditional Variation AutoEncoder (CVAE)} is first used for Conditional Image Generation \cite{DBLP:conf/eccv/YanYSL16}. \citet{DBLP:conf/acl/ZhaoZE17} first proposed Dialogue-CVAE that learned a latent variable to capture discourse-level variations and generated diverse responses by drawing samples from the learned distribution. %This feature naturally suitable for the requirement of UTED task which need select knowledge from specific knowledge distribution.

In UTED, \citet{DBLP:conf/ijcai/SongZCWL19} designed a memory-augmented CVAE architecture (Persona-CVAE) that used a standard memory network to encode persona to a vector representation. In order to capture the relations among dialogue contexts, facts, and responses, \citet{DBLP:journals/corr/abs-1903-09813} introduced a CVAE model that applied three attention variants to model interactions: context-only attention, parallel attention, and context-guided fact attention. 

The CVAE structure is known to suffer the vanishing latent variable problem \cite{DBLP:conf/acl/ZhaoZE17} that the model tends to minimize the KL divergence between prior and posterior to 0, which cause the model can not learn the posterior information. Some methods such as annealing loss trick \cite{DBLP:journals/corr/abs-1903-09813} or bag-of-words loss \cite{DBLP:conf/acl/ZhaoZE17} were applied to alleviate this problem. \citet{DBLP:journals/corr/abs-2003-12738} pointed out that the autoregressive computation of the RNN limited the training efficiency of CVAE, they proposed Variational Transformers models to address the problem.

\textbf{4) Generative Adversarial Networks (GAN)} \cite{DBLP:conf/nips/GoodfellowPMXWOCB14} consists of two alternating training modules: Generator and Discriminator. The training goal of the generator is to generate data similar to the real data in the training set. The Discriminator is usually a binary classification model to determine whether the input is a real training sample or a sample generated by the generator. The training goal is to distinguish the real data and the generated one. In NLP, Adversarial training is usually adopted in Reinforcement Learning \cite{DBLP:conf/emnlp/LiMSJRJ17} framework, the output of the discriminator is used as a reward to train the generative model. The GAN is known to be remarkably difficult to train \cite{DBLP:conf/iclr/ArjovskyB17}, especially in NLP \cite{DBLP:conf/nips/dAutumeMRR19}. However, according to \citet{DBLP:conf/nips/dAutumeMRR19}, GANs do not suffer from exposure bias \footnote{A distributional shift between training data used for learning and model data required for generation.} since the model learns to sample during training.

In UTED, based on adversarial training in dialogue generation \cite{DBLP:conf/emnlp/LiMSJRJ17}, \citet{noauthororeditor} proposed a adversarial approach (Adv) for persona-based dialogue. They used a Seq-to-Seq model as a generator and GRU+MLP as a discriminator. But unlike \citet{DBLP:conf/emnlp/LiMSJRJ17}, the Adv model was strictly adversarial and did not employ pre-training nor auxiliary rewards. \citet{DBLP:conf/aaai/SongZH020} proposed a Reinforcement Learning based Consistent Dialogue Generation approach (RCDG) to exploit the advantages of the natural language inference (NLI) technique to address the inconsistent persona problem. The generator was a PTM (BERT) and the evaluator consisted of two components: an adversarially trained naturalness module and an NLI based consistency module. \citet{DBLP:journals/corr/abs-2007-00067} argued that backward dependency of mutual information was crucial to the variational information maximization lower bound. They proposed Adversarial Mutual Information (AMI) to identify the joint interaction between source and target. In this framework, forward and backward networks could iteratively upgrade or downgrade the instances generated by each other by comparing real and synthetic data distributions. The model used adversarial training to maximize mutual information and minimize data reconstruction space simultaneously.

\textbf{5) Meta-Learning} aims to train the initial parameters of the model so that these parameters can be quickly iterated to new tasks. It is naturally suitable for few-shot or cross-domain tasks. The previous methods learned an iterative function or a learning rule \cite{DBLP:journals/air/VilaltaD02}, while Model-Agnostic Meta-Learning (MAML) proposed by \citet{DBLP:conf/icml/FinnAL17} maximize the sensitivity of the loss function on new tasks. The loss function could be greatly reduced when the parameters were only slightly changed. %For a target task, its dialogue model is obtained by finetuning the initial parameters from MAML with its task-specific training samples.

In UTED, \citet{DBLP:conf/acl/MadottoLWF19} extended MAML to Persona-Agnostic Meta-Learning (PAML). The PAML trained an initial set of parameters that could quickly be adapted to a new persona from a few samples. \citet{DBLP:conf/acl/SongLBYZ20} proposed the Customize Model-Agnostic meta-learning (CMAML) algorithm, which could customize a unique model for each dialogue task. In CMAML, each dialogue model consisted of a shared module, a strobe module, and a private module. The first two modules were shared by all tasks, and the third private module had a unique network structure and parameters to capture the characteristics of the corresponding tasks. They also introduced two steps Customized Model Training method (Private Network Pruning and Joint Meta-learning).%Learning to Customize Model Structures for Few-shot Dialogue Generation Tasks

\textbf{6) Copy mechanism} \cite{DBLP:conf/acl/GuLLL16,DBLP:conf/acl/GulcehreANZB16} uses decoder hidden state to learn whether to copy words or fragments directly from the input or to generate words in the vocabulary. It was widely used in generative models to deal with OOV problem in NLP \cite{DBLP:conf/acl/SeeLM17}. 

In UTED, copy mechanism also played an important role since the decoder needed to copy information from the dialogue context \cite{DBLP:conf/iclr/ZhaoWTX0020}, external knowledge \cite{DBLP:journals/corr/abs-1906-06685,DBLP:journals/corr/abs-1908-09528,DBLP:conf/iclr/KimAK20} or both \cite{DBLP:conf/sigdial/YavuzRCH19,DBLP:conf/acl/LinJHWC20}.

The RefNet \cite{DBLP:journals/corr/abs-1908-06449} model used a decoder switcher to switch between normal generation+copy and a semantic unit copy. The limitation was that the semantic unit copy needs knowledge labeled data such as Holl-E for training. Deepcopy \cite{DBLP:conf/sigdial/YavuzRCH19} designed a decoder copying from multiple knowledge sentences and the dialogue context. It used a feed-forward network to calculate attention score over the context and each knowledge sentence with the decoder’s hidden state. They also introduced a Hierarchical Pointer Network where the decoder hidden state was used to attend over token level representations and the overall fact level representations.

%%%%%%%%%%%%%%%%%%%% LOSS
\subsubsection{Loss Function}
Training objects used in generative UTED models include: %Recent models with separate or multiple training stages \cite{DBLP:conf/acl/SongWZLL20,DBLP:conf/aaai/SongZH020,DBLP:conf/sigir/MengRCSRTR20} often used multiple objective functions or a weighted combination of losses \cite{DBLP:conf/ijcai/SongZCWL19,DBLP:journals/corr/abs-2004-14614,DBLP:conf/ecai/XuLYRRC020}. 

\textbf{1) Language model loss:} Language Models (LM) commonly adopt Cross-Entropy (CE) loss where the final hidden state in the decoder is fed into an output layer with softmax to obtain next token probabilities. These probabilities are then scored using NLL loss where the gold next tokens are taken as labels. For instance, some approach in UTED employed a single Maximum Likelihood Estimation (MLE) loss \cite{DBLP:journals/corr/abs-1911-09728}; some others used multiple losses including LM loss \cite{DBLP:conf/www/ZhaoWHYCW20,DBLP:conf/iclr/ZhaoWTX0020,DBLP:journals/corr/abs-1901-08149,DBLP:conf/aaai/LinXWSLSF20}.

\textbf{2) Knowledge selection loss:} In explicit knowledge selection models, if true labels were given, the knowledge selection loss, which computed the Cross-Entropy loss between gold knowledge distribution and predicted knowledge distribution, was always computed \cite{DBLP:journals/corr/abs-1811-01241,DBLP:conf/iclr/KimAK20}. Training with knowledge selection loss could provide the model with a straightforward signal to direct knowledge selection. If true labels were not given, some word overlap techniques \cite{DBLP:journals/corr/abs-1908-09528,DBLP:journals/access/AhnLP20} were employed to generate pseudo-knowledge labels. \citet{DBLP:conf/ecai/XuLYRRC020} designed Persona Exploration and Exploitation (PEE) framework. The former was based on the VAE structure, the latter used the key-value memory structure and a mutually reinforcing multi-hop memory retrieval mechanism. Apart from the NLL, the PEE model introduced two rule-based persona-oriented loss functions: Persona-oriented Matching (P-Match) loss and Persona-oriented Bag-of-Words (P-BoWs) loss which respectively supervised persona selection in encoder and decoder.

\textbf{3) Priori and posterior loss:} Some losses are introduced to minimizing the gap between training and inference when Teacher-Student or CVAE module is employed. \citet{DBLP:conf/naacl/AroraKR19} proposed a Teacher-Student model called RAM with MLE and mean square loss (MSE). The MSE was used to diminished the gap between the student model and the teacher model. The Persona-CVAE \cite{DBLP:conf/ijcai/SongZCWL19} model was trained with the sum of the following losses: the variational lower bound of the conditional log-likelihood, the cross-entropy loss of persona memory module and type distribution, and the bag-of-words loss \cite{DBLP:conf/acl/ZhaoZE17}. \citet{DBLP:conf/ijcai/LianXWPW19} introduced Posterior Knowledge Selection (PostKS) model and used the NLL, bag-of-words (BOW), and KL-Divergence Loss. The posterior knowledge distribution was used as a pseudo-label for knowledge selection. GLKS \cite{DBLP:journals/corr/abs-1908-09528} used MLE loss, Distant Supervision loss and Maximum Causal Entropy (MCE) loss. DukeNet \cite{DBLP:conf/sigir/MengRCSRTR20} proposed a two phases training scheme: warm-up training phase and dual interaction training phase. In the first phase, MLE losses (knowledge tracking and shifting loss) were used along with the generation loss. In the second phase, KL-loss was used to reduce the impact of inaccurate reward estimation and the overall loss was the combination of MLE losses and KL-loss. Decoupling \cite{DBLP:journals/corr/abs-2004-14614} and COMPAC \cite{DBLP:journals/corr/abs-2010-03205} used NLL for generation and KL-Divergence for minimizing the knowledge selection distribution between the prior and posterior. To train a stronger prior knowledge selection module, \citet{DBLP:conf/emnlp/ChenMLCXXZ20} used the response in BOW format as the posterior information and an NLL loss over the vocabulary in a Teacher-Student framework. Meanwhile, a "fix" operation was introduced to ensure that the prior knowledge distribution did not affect the posterior selection.

\textbf{4) Mutual information loss:} Mutual information between dialogue context and response can be used to improve diversity \cite{DBLP:conf/naacl/LiGBGD16}. In UTED, \citet{DBLP:conf/conll/ZemlyanskiyS18} introduced a DiscoveryScore Metric, which was based on maximizing the mutual information between the current dialogue and the revealed knowledge. \citet{DBLP:conf/acl/DuB19} designed an iterative training process and integration method based on Boosting, combining this method with different training and decoding paradigms as a basic model, including mutual information-based decoding and reward-enhanced maximum likelihood learning. The reward enhancement was to add exponential level feedback similar to the evaluation metric in MLE. ZRKGC \cite{DBLP:journals/corr/abs-2008-12918} used Mutual Information Loss between response and grounding rate of latent knowledge. Besides, marginal log-likelihood with Generalized EM method \cite{DBLP:books/lib/Bishop07} was used and Knowledge Selection Loss was adopted. 
%%%%%%%%%%%%%%%%%%%%%%%%%%%%%%%%%%%%%%%%%%%%%%%%%%%%%%Specifically, the metric was used for re-ranking response candidates after beam search. 
%\citet{DBLP:journals/csl/Tam20} proposed a Cluster-based Beam Search (CBS) that used K-means to cluster hypotheses into K semantically similar groups at every decoding step. %\citet{DBLP:conf/naacl/DuB19} proposed Lexicalized Probabilistic Context-Free Grammar (L-PCFG) to use syntactic tree structures and lexicalized grammar. The syntactic tree concluded vertices, edges, and bijection. Each vertex comprised lexicon, label, production rule, and index of its heir. L-PCFG used different LSTM to encode lexicalization and syntactic history and decoded each tree node with the guidance of both rules and words.

\textbf{5) Other losses:} When different labels are given, auxiliary loss \cite{DBLP:conf/emnlp/MajumderLNM20,DBLP:conf/emnlp/ScialomTSG20} can be leveraged in a multi-task learning schema. L-PCFG \cite{DBLP:conf/naacl/DuB19} used both syntactic tree structures and lexicalized grammar and was trained by minimizing the NLL of both rules and words. RefNet \cite{DBLP:journals/corr/abs-1908-06449} used generation loss, reference loss and switcher loss (switching between Generation decoder and Reference Decoder). In a similar way, \citet{DBLP:journals/corr/abs-2005-03174} proposed a convergent and divergent decoding strategy that could generate informative and diverse responses considering not only given inputs (context and facts) but also inputs-related topics. They used NLL, copy loss, and switcher loss (selecting convergent or divergent decoding strategy). \citet{DBLP:conf/acl/LiRKWBCW20} argued that models trained with maximum likelihood estimation (MLE): (i) rely too much on copying from the context, (ii) contain repetitions within utterances, (iii) overuse frequent words, and (iv) at a deeper level, contain logical flaws. To address these problems, they used the idea of Unlikelihood \cite{DBLP:conf/iclr/WelleckKRDCW20} technique to construct four additional losses besides MLE in the UTED task.

In Table \ref{Generative}, we summarize the current generation-based models using the categories we introduced in this chapter. 

\begin{table}[h]
\footnotesize
\caption{Module comparison between different Generative models. "PTM" means Pre-trained Model. "Supp." means KS with Supplementary information. "EarI." stands for KS with Early Interaction.}
\label{Generative}
\begin{tabular}{l|ccc}
\toprule
Models (Papers)  & Knowledge \& Dialogue Encoding & Knowledge Selection & Response Generation  \\
\midrule
KGNCM \cite{DBLP:conf/aaai/GhazvininejadBC18}  & w/o pre-training (GRU) & Implicit (EarI.) & Seq-to-Seq \\ %ghaz
CaKe \cite{DBLP:journals/corr/abs-1906-06685}, GLKS \cite{DBLP:journals/corr/abs-1908-09528} & w/o pre-training (GRU) &Implicit (EarI.) & Seq-to-Seq, Copy \\
Persona-CVAE \cite{DBLP:conf/ijcai/SongZCWL19}&w/o pre-training (GRU)  & Implicit & Seq-to-Seq, CVAE \\%Exploiting Persona Information for Diverse Generation of Conversational Responses 
KIC \cite{DBLP:conf/acl/LinJHWC20}       & w/o pre-training (LSTM) & Implicit & Seq-to-Seq, Copy \\%Knowledge-Interaction and Knowledge-Copy
RefNet \cite{DBLP:journals/corr/abs-1908-06449} & w/o pre-training (LSTM) &Implicit (EarI.) & Seq-to-Seq, Copy \\
L-PCFG \cite{DBLP:conf/naacl/DuB19} & w/o pre-training (LSTM) & Implicit & Seq-to-Seq, Syntactic Tree \\%Lexicalized Probabilistic Context-Free Grammar %concate persona to context
Deepcopy \cite{DBLP:conf/sigdial/YavuzRCH19} & w/o pre-training (LSTM) & Implicit & Seq-to-Seq, Copy \\ %Deepcopy: Grounded response generation with hierarchical pointer networks
TED \cite{DBLP:conf/cikm/ZhengZ19}, GDR \cite{DBLP:conf/acl/SongWZLL20},& w/o pre-training (Transformer) &Implicit & Seq-to-Seq \\%Enhancing Converstional Dialogue models with grounded knowledge % generate delete rewrite
 Interleave \cite{DBLP:journals/corr/abs-1911-09728}&  \\%Improving Conditioning in Context-Aware Sequence
ITDD \cite{DBLP:conf/acl/LiNMFLZ19}, CAT \cite{DBLP:conf/emnlp/MaZS020} &w/o pre-training (Transformer) &Implicit (EarI.) & Seq-to-Seq \\
CG+CVAE \cite{DBLP:journals/corr/abs-1903-09813}& re-trained GloVe (GRU)  & Implicit & Seq-to-Seq, CVAE \\ %Knowledge-Grounded Response Generation with Deep Attentional Latent-Variable
CDD \cite{DBLP:journals/corr/abs-2005-03174} & GloVe (GRU) & Implicit &Seq-to-Seq, Copy \\ %Fact-based Dialogue Generation with Convergent and Divergent Decoding
Adv \cite{noauthororeditor} & GloVe (GRU) & Implicit & Seq-to-Seq, GAN \\%Adversarial Approach to Persona Based Dialogue Agents
RAM \cite{DBLP:journals/corr/abs-2005-06128} & GloVe (LSTM) & Implicit (EarI.) &Seq-to-Seq \\% Response-Anticipate
CMAML \cite{DBLP:conf/acl/SongLBYZ20} & GloVe (LSTM) & Implicit & Seq-to-Seq, Meta-Learning \\%Learning to Customize Model Structures for Few-shot Dialogue
CBS \cite{DBLP:journals/csl/Tam20} & GloVe (LSTM) & Implicit & Seq-to-Seq, Copy \\%cluster-based beam search  
PAML \cite{DBLP:conf/acl/MadottoLWF19} & GloVe (Transformer) & Implicit &Seq-to-Seq, Meta-Learning \\%Personalizing dialogue agents via meta-learning
DRD \cite{DBLP:conf/iclr/ZhaoWTX0020} & PTM (Transformer) & Implicit & Seq-to-Seq, Copy \\%Low-resource knowledge-gournded
KIF \cite{DBLP:journals/corr/abs-2004-12744}, Unlikelihood \cite{DBLP:conf/acl/LiRKWBCW20} & PTM (Transformer) & Implicit &  Seq-to-Seq \\ %KNN-based \
Multi-Input \cite{DBLP:conf/acl/GolovanovKNTTW19}, LIC \cite{golovanov2020lost}, & PTM (GPT-1) & Implicit &  Seq-to-Seq \\  %concate K to C %Large-scale transfer learning for natural language generation
MKST \cite{DBLP:conf/www/ZhaoWHYCW20}, $\mathcal{P}^2$ Bot \cite{DBLP:conf/acl/LiuCCLCZZ20},  &  \\%Multiple Knowledge Syncretic Transformer%Lost in conversation %You impress me: Dialogue generation via mutual persona perception
Decoupling \cite{DBLP:journals/corr/abs-2004-14614} &  \\%Unsupervised Injection of Knowledge  
TransferTransfo \cite{DBLP:journals/corr/abs-1901-08149} & PTM (GPT-1) & Implicit & Seq-to-Seq, Copy \\
SSS \cite{DBLP:journals/corr/abs-2005-14315} & PTM (ELMo/BERT)& Implicit & Seq-to-Seq, Copy\\%On Incorporating Structural Information
RCDG \cite{DBLP:conf/aaai/SongZH020}&PTM (BERT) & Implicit & Seq-to-Seq, GAN\\%Generating Persona Consistent Dialogues by Exploiting Natural Language Inference
Adapter-Bot \cite{DBLP:journals/corr/abs-2008-12579}  &PTM (DialoGPT) & Implicit &Seq-to-Seq \\%The Adapter-Bot All-In-One Controllable Conversational Model
CGRG \cite{DBLP:journals/corr/abs-2005-00613} &PTM (GPT-2) & Implicit  &  Seq-to-Seq \\%A Controllable Model of Grounded Response Generation
ZRKGC \cite{DBLP:journals/corr/abs-2008-12918} &PTM (UNILM) & Implicit & Seq-to-Seq\\%zero-resource
\midrule
GE \cite{DBLP:conf/acl/BaoHWLW19} & w/o pre-training (GRU) & Explicit & Seq-to-Seq, RL\\%Know More about Each Other: Evolving Dialogue Strategy via Compound Assessment
Match-Reduce \cite{DBLP:journals/access/AhnLP20} & w/o pre-training (Transformer) & Explicit & Seq-to-Seq\\%Exploiting Text Matching Techniques for knowledge-grounded
PostKS \cite{DBLP:conf/ijcai/LianXWPW19}  & GloVe (GRU) & Explicit & Seq-to-Seq\\
DiffKS \cite{DBLP:conf/emnlp/ZhengCJH20}  & GloVe (GRU) & Explicit (Supp.) & Seq-to-Seq, Copy\\
AKGCM \cite{DBLP:journals/corr/abs-1903-10245} & GloVe (LSTM) & Explicit (Supp.) & Seq-to-Seq, RL, Copy\\%Knowledge Aware Conversation Generation with Reasoning on Augmented
RR \cite{DBLP:conf/emnlp/WestonDM18} &GloVe (LSTM) & Explicit & Seq-to-Seq \\%Retrieve an Refine
TMN \cite{DBLP:journals/corr/abs-1811-01241}, TF \cite{DBLP:conf/interspeech/GopalakrishnanH19}    & PTM (Transformer) & Explicit & Seq-to-Seq \\
%wow %Topical chat
KnowledGPT \cite{DBLP:conf/emnlp/ZhaoWXTZY20} &PTM (GPT-2/BERT)&Explicit &Seq-to-Seq\\%RL for KS
SKT \cite{DBLP:conf/iclr/KimAK20}, DukeNet \cite{DBLP:conf/sigir/MengRCSRTR20},   & PTM (BERT) & Explicit (Supp.) & Seq-to-Seq, Copy \\%A Dual Knowledge Interaction Network for Knowledge-Grounded Conversation
PIPM+KDBTS \cite{DBLP:conf/emnlp/ChenMLCXXZ20} &&&\\
PD-NRG \cite{DBLP:journals/corr/abs-2005-12529} & PTM (GPT-1) & Explicit & Seq-to-Seq \\ %policy-Driven Topical chat
%COMPAC \cite{DBLP:journals/corr/abs-2010-03205} & PTM (GPT-2) & Explicit (Supp.) & Seq-to-Seq\\%Like hiking?
\bottomrule
\end{tabular}
\end{table}%TF+AMI random initialized LSTM%GDR random initialized Transformer %PEE glove GRU
%AMI: https://github.com/ ZJULearning/AMI

%%%%%%%%%%%%%%%%%%%%%%%%%%%%
%%%%%%%%%%%%%%%%%%%%%%%%%%%%  Evaluation
%%%%%%%%%%%%%%%%%%%%%%%%%%%%
\section{Evaluation Metrics}
At present, the evaluation methods of the UTEDS can be divided into two categories: auto evaluation and human evaluation. \textbf{Auto Evaluation} can be calculated by computer and is often based on accuracy\cite{DBLP:journals/corr/abs-1811-01241}, word overlaps \cite{DBLP:conf/acl/PapineniRWZ02}, word embedding similarity \cite{DBLP:conf/emnlp/LiuLSNCP16} between the generated response and the ground-truth response. Most recently, model-based \cite{DBLP:conf/iclr/ZhangKWWA20} auto evaluation is introduced. \textbf{Human Evaluation} needs humans to evaluate the quality of responses generated by models. Auto evaluation and human evaluation are normally used together to reflect the models' performance.

\subsection{Auto Evaluation}
In this section, we introduce the automatic evaluation metrics currently used in the UTEDS. 

\subsubsection{Retrieval-based}

The ranking of candidate answers is the core of the retrieval DS. \citet{DBLP:conf/ijcai/ZhaoTWX0Y19} used R@N which evaluates whether the correct candidate is retrieved in the top N results. \citet{DBLP:journals/corr/abs-1811-01241} adopted Hit@1 (also called Recall@1) which calculated the accuracy of selecting the right knowledge.

\subsubsection{Generative-based}

\textbf{Probability-based:} %%%%%%%%%%%%%%%%%%%%%%%%%%%%%%%%%%%%%%%%%%need to clarify bigger better or smaller better
\textit{Perplexity (PPL) \cite{DBLP:conf/nips/BengioDV00}} is usually employed to measure the probability of the occurrence of a sentence in LM. While in the DS, the PPL measures how well the model predicts a response, a lower perplexity score indicates better generation performance. Other than target words, some researchers \cite{DBLP:conf/emnlp/MajumderLNM20} used Byte Pair Encoder (BPE) \cite{DBLP:journals/corr/abs-2004-12744,DBLP:conf/interspeech/GopalakrishnanH19} to compute perplexity. The BPE perplexity can better address the OOV problem. \textit{Entropy-n \cite{DBLP:conf/nips/ZhangGGGLBD18}} reflects how evenly the empirical n-gram distribution is for a given sentence. 

\textbf{Word overlap with ground truth:} 
\textit{BLEU-n \cite{DBLP:conf/acl/PapineniRWZ02}} measures the n-gram overlap between the generated response and the golden response. \textit{BLEU} calculates the geometric average of the accuracy of BLEU-n. \textit{ROUGE-n \cite{lin2004rouge}} is based on the calculation of the recall rate of the common sub-sequence of generating response and the real one. \textit{METEOR \cite{DBLP:conf/wmt/LavieA07}} further considers the alignment between the generated and the real responses to improve BLEU. WordNet is adopted to calculate the matching relationship among specific sequence matching, synonym, root, interpretation, etc. \textit{NIST \cite{doddington2002automatic}} is an improvement of BLEU by summing up each weighted co-occurrence n-gram segments, then dividing it by the total number of n-gram segments. \textit{Distinct-n \cite{DBLP:conf/naacl/LiGBGD16}} measures the diversity of reply by calculating the proportion of distinct n-grams in the total number of n-grams to evaluate the diversity of generated responses. \textit{Word-level F1 \cite{DBLP:conf/emnlp/RajpurkarZLL16}} treats the prediction and ground truth as bags of tokens and measures the average overlap between the prediction and ground truth answer. \textit{Exact Match} (EM) \cite{DBLP:journals/corr/abs-1902-00821} requires the answers to have an exact string match with human-annotated answer spans. \citet{DBLP:journals/corr/abs-1911-09728} proposed \textit{context use percentage} to measure how well the model utilizes the relevant information from the context. \citet{DBLP:conf/acl/LiRKWBCW20} measured the fraction of generated n-grams that appear in the original context as \textit{context repetition}.

\textbf{Embedding-based:}
\textit{Greedy Matching \cite{DBLP:conf/bea/RusL12}} is an embedding-based metric that greedily matches each word in the generated sequence to a reference word based on the cosine similarity of their embeddings. The final score is then an average over all the words in the generated sequence. \textit{Embedding Average \cite{DBLP:journals/corr/WietingBGL15a}} computes a sentence embedding for both the generated sequence and the ground-truth response by taking an average of word embeddings. The score is then a cosine similarity of the average embedding for both the generated and reference sequence. \textit{Vector Extrema \cite{forgues2014bootstrapping}} follows a similar setup to Embedding Average, where the score is the cosine similarity between sentence embeddings. Rather than taking an average over word embeddings, this method identifies the maximum value for each dimension of the word embedding. Taking the maximum is motivated by the idea that common words will be de-emphasized as they will be closer to the origin. Vector Extrema showed some advantages on dialogue tasks \cite{DBLP:conf/emnlp/LiuLSNCP16}. \textit{Skip-Thought} \cite{DBLP:conf/nips/KirosZSZUTF15} uses a recurrent neural network to produce a sentence-level embedding for the generated and reference sequences. Cosine similarity is then computed between the two embeddings. \citet{DBLP:conf/sigdial/YavuzRCH19} employed \textit{CIDEr} \cite{DBLP:conf/cvpr/VedantamZP15} to measure the cosine similarity of generated sentence and reference based-on TF-IDF vector. To train the diversity of responses, \citet{DBLP:conf/naacl/DuB19} used a \textit{K-means clustering} on the sentence embeddings of responses and calculated average Squared Euclidean Distance between members of each cluster.

\textbf{Model-based:}
\textit{BERTScore \cite{DBLP:conf/iclr/ZhangKWWA20}} uses a pre-trained BERT model to greedily match each word in a reference response with one word in the generated sequence. By doing so, it computes the recall of the generated sequence. BERTScore was shown to have a strong system-level and segment-level correlation with human judgment on several machine translation and captioning tasks. In order to approximate manually labeled ratings, \citet{DBLP:conf/acl/MehriE20} proposed \textit{UnSupervised and Reference-free (USR)} evaluation metric. From the perspective of turn level, The USR used five sub-evaluation metrics (Understandable/Natural/Maintaining Context/Interesting/Using Knowledge) and a general evaluation metric (Overall Quality). A regression model was used on top of a Masked Language Modelling (MLM) and trained to reproduce the overall score from each of the specific quality scores rated by annotators. The \textit{MLM} was a fine-tuned RoBERTa \cite{DBLP:journals/corr/abs-1907-11692} model. The likelihood of a response estimated by the fine-tuned RoBERTa model is used as an automatic metric for evaluating the understandability and naturalness of responses. 

\textbf{Knowledge based:}
\citet{DBLP:conf/acl/QinGBLGDCG19} calculated F1 with non-stop word tokens in the response that are present in the document but not present in the context. \textit{Knowledge R/P/F1} \cite{DBLP:journals/corr/abs-1811-01241} measures the F1 overlap of the model’s output with the external knowledge. \citet{DBLP:conf/ijcai/SongZCWL19} proposed \textit{Persona Coverage} to evaluate how well persona information was leveraged. \citet{DBLP:conf/emnlp/MaZS020} proposed Knowledge-Utilization (KU) to measure how many N-grams in external knowledge are used in responses. They also introduced Quality of Knowledge Utilization (QKU) to measure the quality of the N-grams. \textit{Point-wise mutual information (PMI) \cite{DBLP:conf/acl/ChurchH89}} aims to evaluate how smoothly the generated responses are related to the context and knowledge. \citet{DBLP:conf/acl/MadottoLWF19} and \citet{DBLP:conf/acl/SongWZLL20} trained NLI models to measure the persona consistency between persona sentences and generated responses, their metrics were named \textit{C score} and \textit{Ent}, respectively.

\textbf{Other:}
\textit{Average length} of the generated response is often used \cite{DBLP:conf/acl/QinGBLGDCG19}, a response with more words is usually considered to contain more information.

\subsection{Human Evaluation}%a new method is evaluate increasingly by turns
Similar to the Ranking method of the retrieval models, we can classify human evaluations into 2 categories: \textbf{Point-wise Grading} and \textbf{Pair-wise Comparison}. In Point-wise Grading, workers need to rate each model independently. In Pair-wise Comparison, workers are required to compare different models and choose the preferred one. Meanwhile, according to different scoring objects, we can further divide human evaluations into 2 other categories: \textbf{Turn-level} and \textbf{Dialogue-level}. Turn-level evaluation scores the quality of the response and is an offline procedure. Dialogue-level evaluation assesses the quality of multi-turn responses or the entire conversation through interaction with the model. The same person usually plays the role of both the user (one who interacts with the system) and the evaluator \cite{DBLP:conf/sigdial/FinchC20}. For dialogue quality, Appropriateness \cite{DBLP:conf/ijcai/ZhouYHZXZ18}, Coherence \cite{DBLP:journals/corr/abs-2008-12918}, Engagingness \cite{DBLP:conf/sigir/MengRCSRTR20}, Fluency \cite{DBLP:conf/acl/LinJHWC20}, Interest \cite{DBLP:conf/interspeech/GopalakrishnanH19}, and Naturalness \cite{DBLP:journals/corr/abs-1908-06449} are often used in human evaluations. For the utilization of external knowledge, Informativeness \cite{DBLP:conf/acl/LinJHWC20}, and Relevance \cite{DBLP:conf/www/ZhaoWHYCW20} are usually adopted. In Table \ref{human evaluation}, We summarize human evaluation metrics in current UTED models. 

%turn level - dialogue level / pair wise compare - point wise rate
\begin{table}[h]
\footnotesize
\caption{Models with different manual evaluation methods. "Point / Pair / Turn / Dialog" are short for "Point-wise Grading / Pair-wise Comparison / Turn-level / Dialogue-level", respectively.}
\label{human evaluation}
\begin{tabular}{l|cc}
\toprule
Models  & Dataset & Categories   \\

\midrule
TMN \cite{DBLP:journals/corr/abs-1811-01241}     & WoW &  Point, Dialog\\%chat with models and rate the conversations
KIC \cite{DBLP:conf/acl/LinJHWC20}      & WoW &  Point, Turn \\%Fluency / Coherence / Informativeness
KIF \cite{DBLP:journals/corr/abs-2004-12744}  & WoW & Pair, Dialog \cite{DBLP:journals/corr/abs-1909-03087} \\%Acute-Eval dialogue evaluation system 
DiffKS \cite{DBLP:conf/emnlp/ZhengCJH20} &WoW &Pair, Turn / Point, Dialog\\%Naturalness, Appropriateness
PostKS \cite{DBLP:conf/ijcai/LianXWPW19} & WoW, Persona-Chat  &  Point, Dialog \\%the overall quality
MKST \cite{DBLP:conf/www/ZhaoWHYCW20} & WoW, Persona-Chat & Point, Turn \\%appropriateness \cite{DBLP:conf/ijcai/ZhouYHZXZ18}, knowledge relevance\cite{DBLP:conf/acl/LiuFCRYL18}
%Decoupling \cite{DBLP:journals/corr/abs-2004-14614}     & WoW, Persona-Chat, LIGHT & - \\
%Adapter-Bot \cite{DBLP:journals/corr/abs-2008-12579}   & WoW, Persona-Chat, ED & - \\
AKGCM \cite{DBLP:journals/corr/abs-1903-10245} & WoW, Holl-E & Pair, Turn \\%appropriateness / informativeness (compare different models)
SKT \cite{DBLP:conf/iclr/KimAK20} & WoW, Holl-E & Point, Dialog \cite{DBLP:journals/corr/abs-1811-01241}\\%chat with models and rate the conversations
DukeNet \cite{DBLP:conf/sigir/MengRCSRTR20}, PIPM+KDBTS \cite{DBLP:conf/emnlp/ChenMLCXXZ20} & WoW, Holl-E & Point, Turn  \\%Appropriateness / Informativeness / Engagingness 
%DRD \cite{DBLP:conf/iclr/ZhaoWTX0020} & WoW, CMU\_DoG & - \\
Unlikelihood \cite{DBLP:conf/acl/LiRKWBCW20} & WoW, ConvAI2 & Pair, Turn \\%pair wise readability and accuracy of the respons
Match-Reduce \cite{DBLP:journals/access/AhnLP20}, KnowledGPT \cite{DBLP:conf/emnlp/ZhaoWXTZY20} & WoW, CMU\_DoG & Point, Turn \\%Appropriateness / Informative gaine\cite{DBLP:conf/acl/FanJPGWA19}
ZRKGC \cite{DBLP:journals/corr/abs-2008-12918} & WoW, CMU\_DoG, Topical-Chat & Point, Turn \\%fluency / context coherence / knowledge relevance
ITDD \cite{DBLP:conf/acl/LiNMFLZ19}, CAT \cite{DBLP:conf/emnlp/MaZS020} &CMU\_DoG&Point, Turn\\%Fluency, Coherence, and Informativeness
RefNet \cite{DBLP:journals/corr/abs-1908-06449}, GLKS \cite{DBLP:journals/corr/abs-1908-09528} & Holl-E & Point, Turn  \\%Naturalness / Informativeness / Appropriateness / Humanness
TF \cite{DBLP:conf/interspeech/GopalakrishnanH19} & Topical-Chat & Point, Turn \\%comprehensible / on-topic / interesting
PD-NRG \cite{DBLP:journals/corr/abs-2005-12529} & Topical-Chat & Pair, Turn \\ %appropriate (compare different models)
CMR \cite{DBLP:conf/acl/QinGBLGDCG19} & CbR & Pair, Turn \\%Relevance and Informativeness (compare different models)
RAM \cite{DBLP:journals/corr/abs-2005-06128}, CGRG \cite{DBLP:journals/corr/abs-2005-00613}& CbR & Point, Turn\\% overall quality , relevance with documents , and informativeness  %relevance / background consistency
\midrule
LIC \cite{golovanov2020lost}, PAML \cite{DBLP:conf/acl/MadottoLWF19}, CMAML \cite{DBLP:conf/acl/SongLBYZ20}, GDR \cite{DBLP:conf/acl/SongWZLL20},& Persona-Chat & Point, Turn\\%aggregate fluency score, consistency and engagement to get dialogue quality / Persona Detection %fluency / consistency  %Quality / Task Consistency (with the characteristics of a certain task)
Per-CVAE \cite{DBLP:conf/ijcai/SongZCWL19}, PEE \cite{DBLP:conf/ecai/XuLYRRC020},  RCDG \cite{DBLP:conf/aaai/SongZH020} &  \\% Engagingness / Variety / Persona Detection %Fluency / Engagingness / Consistency / Persona Detection %consistent / fluency / relevance / informativeness%effectiveness of the model-based evaluation / dialogue quality
\midrule
DiscoveryScore \cite{DBLP:conf/conll/ZemlyanskiyS18}, $\mathcal{P}^2$ Bot \cite{DBLP:conf/acl/LiuCCLCZZ20}, & Persona-Chat & Point, Dialog \\
RAML \cite{DBLP:conf/nips/NorouziBCJSWS16}+Boosting \cite{DBLP:conf/acl/DuB19}& \\ %Fluency / Engagingness / Consistency (rate after chatting with model) %appropriate (compare different models) %Dialogue quality 
\midrule
TF+AMI \cite{DBLP:journals/corr/abs-2007-00067}, COMPAC \cite{DBLP:journals/corr/abs-2010-03205}  &Persona-Chat & Pair, Turn \\%dialogue quality %Fluency / Engagement / relevance /Persona Detection
GE \cite{DBLP:conf/acl/BaoHWLW19}&Persona-Chat &Pair, Dialog\\%Overall / Coverage / Concise / Coherence 
RR \cite{DBLP:conf/emnlp/WestonDM18} & Persona-Chat & Point and Pair, Dialog \\% engagingness, consistency, fluency
TransferTransfo \cite{DBLP:journals/corr/abs-1901-08149}, Deepcopy \cite{DBLP:conf/sigdial/YavuzRCH19} &ConvAI2 & Point, Turn \\% fluency, consistency, engagingness,  %Appropriateness, Fact inclusion
CG+CVAE \cite{DBLP:journals/corr/abs-1903-09813}, CBS \cite{DBLP:journals/csl/Tam20}, CDD \cite{DBLP:journals/corr/abs-2005-03174}& DSTC-7 & Point, Turn \\%Context  Relevance / Interest / Fluency / Knowledge Relatedness %relevance and appropriateness” and “interest and informativeness % Appropriateness Informativeness Fluency 
Moel \cite{DBLP:conf/emnlp/LinMSXF19} & ED & Point, Turn / Pair, Dialog\\%human(Empathy/Sympathy,Relevance,Fluency) / human (Compare)
\bottomrule
\end{tabular}
\end{table}

%\citet{DBLP:journals/corr/abs-2004-14614} conducted the evaluation considering the \textit{knowledge gap}. Specifically, the metrics are drawn as curves with regard to varying amounts of external knowledge. The curves show a more comprehensive evaluation of the models, clarifying how effectively the models utilize external knowledge. RCDG \cite{DBLP:conf/aaai/SongZH020} exploited human evaluations to verify the effectiveness of the model-based evaluation. Responses from all models were divided into three categories and response-persona pairs were randomly sampled from each category. The judges were instructed to give a 5-scale score to each pair, the judges evaluated samples from each category, rather than from each model. 

There are some other interesting human evaluation metrics recently introduced. \citet{DBLP:conf/sigdial/MehriE20} introduced 18 metrics from the perspective of each turn or whole dialogue. KIF \cite{DBLP:journals/corr/abs-2004-12744} employed Acute-Eval dialogue evaluation system \cite{DBLP:journals/corr/abs-1909-03087} which comparing two full dialogues.

%%%%%%%%%%%%%%%%%%%%%%%%%%%%%%%%%%%%%%%%%%%%%%%%%%%%%%%%%%%
\section{Analysis}

In Table \ref{ARW}, \ref{HollE}, \ref{WoW} and \ref{Persona}, we present some auto evaluation results of the current UTED models from original papers. We did not compare human evaluations in different articles because they lacked a unified standard. It should be pointed out that these experimental results can only be used as a reference due to the differences in data processing scheme \cite{DBLP:conf/iclr/KimAK20} and experimental settings. For example, generative models might use different data processing schema on the same dataset \cite{DBLP:conf/iclr/KimAK20}; retrieval models on ARW, Persona-Chat, and WoW datasets used 10, 20, and 100 candidates, respectively. In this section, we analyze these experimental results and present some valuable conclusions.

\subsection{Retrieval Models}

\begin{table}[h]
\footnotesize
\caption{Experimental results of Retrieval models. For WoW, we use the Test Seen and Predicted Knowledge version. For Persona-Chat, we use the original persona version. F1 is the unigram overlap of the model’s prediction with the golden response. * means the results are R@1/3/10.}
\label{ARW}
\begin{tabular}{lcc|lccc}
\toprule
Models  & Dataset & R@1/2/5 \% &Models  & Dataset & R@1/2/5 \% & F1  \% \\
\midrule
TF-IDF \cite{lowe2015incorporating}  & ARW & 41.0/54.5/70.8 &IR baseline \cite{DBLP:journals/corr/abs-1811-01241} &WoW & 17.8/--- ---/--- --- & 12.7\\%10 candidates 
KE \cite{lowe2015incorporating}     & ARW & 41.3/55.4/82.4 & BoW MN \cite{DBLP:journals/corr/abs-1811-01241}     &WoW & 71.3/--- ---/--- --- & 15.6 \\%100 cand
TF-IDF \cite{DBLP:conf/acl/KielaWZDUS18}     & Persona-Chat & 41.0/--- ---/--- --- &    TMN \cite{DBLP:journals/corr/abs-1811-01241}          &WoW & 87.4/--- ---/--- --- & 15.4\\%tf-idf weighted cosine similarity (cosine is also a interaction)
Starspace \cite{DBLP:conf/acl/KielaWZDUS18}       & Persona-Chat & 49.1/--- ---/--- --- &     DGMN \cite{DBLP:conf/ijcai/ZhaoTWX0Y19}  & CMUDoG        &   65.6/78.3/91.2  &  \\%K concat to C 
KV-Profile Mem \cite{DBLP:conf/acl/KielaWZDUS18}  & Persona-Chat  & 51.1/--- ---/--- ---  &  FIRE \cite{DBLP:journals/corr/abs-2004-14550}  & CMUDoG & 81.8/90.8/97.4  & \\%K concat to C %HAKIM 19 cand
HAKIM \cite{DBLP:conf/aaai/Sun0XYX20}     & Persona-Chat  & 57.6/72.9/89.9   &   HAKIM \cite{DBLP:conf/aaai/Sun0XYX20}     &  CMUDoG       &   82.7/93.8/99.5 & \\
DGMN \cite{DBLP:conf/ijcai/ZhaoTWX0Y19}  &Persona-Chat  & 67.6/80.2/92.9    &  LSTM \cite{DBLP:conf/emnlp/MazareHRB18}* & M-Persona & 66.3/79.5/90.6 &\\
FIRE \cite{DBLP:journals/corr/abs-2004-14550}  & Persona-Chat & 81.6/91.2/97.8  & Transformer \cite{DBLP:conf/emnlp/MazareHRB18}* & M-Persona & 74.4/85.6/94.2 & \\

\bottomrule
\end{tabular}
\end{table}

In Table \ref{ARW}, we present the results of retrieval-based models. For an earlier ARW task \cite{lowe2015incorporating}, the TF-IDF weighted cosine similarity did not use external knowledge. The Knowledge Encoder (KE) model used the external knowledge, but its performance was restrained by a regular RNN encoder and was only slightly better than TF-IDF, which indicated that simple encoders were not able to capture enough semantic information in UTED task.

%Besides the ARW task, models with external information were always better. Hence we do not present the results of the model without external knowledge. 

For Persona-Chat \cite{DBLP:conf/acl/KielaWZDUS18}, Starspace learned the similarity between the context and the response by optimizing the embedding directly with the margin ranking loss and k-negative sampling, KV Profile Memory used an attention-based key-value memory network to select a response. TF-IDF/Starspace/KV Profile Memory models served as the baselines which concatenated the profiles with dialogue context and their performance was outperformed by the following models that had more complicated structures and encoded context, knowledge, and response separately.

For both Persona-Chat and CMUDoG, HAKIM \cite{DBLP:conf/aaai/Sun0XYX20} employed a representation-based fusion method and learned an interaction vector of context and knowledge. DGMN \cite{DBLP:conf/ijcai/ZhaoTWX0Y19} adopted the Interaction-based fusion method and performed a shallow matching against all fusion matrices. FIRE \cite{DBLP:journals/corr/abs-2004-14550} also used Interaction-based Fusion but with a multi-levels Deep Matching. FIRE was better than DGMN on both Persona-Chat and CMUDoG, which indicated the advantage of multi-levels Deep Matching. HAKIM was worse than the other two on Persona-chat while better than them on CMUDoG. The reason might lie in the Fusion method and the difference between datasets. HAKIM employed Representation-based Fusion while others used Interaction-based. CUMDoG has a longer average length of utterances and external knowledge sentences than Persona-Chat, this entails a data sparsity problem that the dialogue context only contains little external knowledge. %The Representation-base Fusion and sequential interaction of HAKIM can better address this data sparsity by learning a semantic vector than the Interaction-based methods. %Since all 3 models used glove embeddings except FIRE used two additional embeddings.

For WoW \cite{DBLP:journals/corr/abs-1811-01241}, IR baseline used a simple word-overlap method and got the worst R@1 and F1. BoW MN was a Bag-of-Words Memory Network \cite{DBLP:conf/nips/SukhbaatarSWF15}, TMN was a pre-trained Transformer based Memory Network and fine-tuned on WoW. The latter benefits from the representation ability of the Transformer structure and had better results. For X-Persona \cite{DBLP:journals/corr/abs-2003-07568}, the Transformer-based model was also better than the LSTM-based one.

\subsection{Generative Models}

%F1 is calculated by https://github. com/facebookresearch/ParlAI/blob/master/parlai/core/metrics.py

\begin{table}[h]
\footnotesize
\caption{The Generative models' performance on Holl-E/CMUDoG/Topical-Chat/CbR datasets. We only show the Single-reference version of Holl-E and the Rare version of Topical-Chat. * means different data processing schema. \# means div-1/2 from \citet{DBLP:conf/aaai/GhazvininejadBC18}. "R1" stands for Recall@1 and is for knowledge selection.} %\% means the values are multiplied by 100.\^ means the F1 is from \citet{DBLP:conf/acl/QinGBLGDCG19}}
\label{HollE}
\begin{tabular}{l|c|c|c|c|c|c|c}

\toprule
Model & Dataset & F1\% & BLEU(BLEU-1/2/3/4)\% & ROUGE(1/2/L)\% & Distinct(1/2) & PPL & R1\%\\
\midrule
GTTP \cite{DBLP:conf/emnlp/MogheABK18} & Holl-E & &--- ---(--- ---/--- ---/--- ---/ 8.73)& 23.20/ 9.91/17.35&&&\\ 

%BiDAF \cite{DBLP:conf/emnlp/MogheABK18} & Holl-E & 39.69 & 28.85(--- ---/--- ---/--- ---/--- ---) &39.68/33.72/35.91 & &  &\\

SSS(BERT) \cite{DBLP:journals/corr/abs-2005-14315} & Holl-E &  & --- ---(--- ---/--- ---/--- ---/22.78)  &40.09/27.83/35.20 & &  &\\

CaKe \cite{DBLP:journals/corr/abs-1906-06685} & Holl-E & & 26.02(--- ---/--- ---/--- ---/--- ---) & 42.82/30.37/37.48 & & &\\

RefNet \cite{DBLP:journals/corr/abs-1908-06449} & Holl-E & 40.18 & 27.00(--- ---/--- ---/--- ---/--- ---) & 42.87/30.73/37.11 & & &\\

GLKS \cite{DBLP:journals/corr/abs-1908-09528} & Holl-E & & & 43.75/31.54/38.69 & &  &\\

SKT \cite{DBLP:conf/iclr/KimAK20} & Holl-E* &29.8&   &   & & 48.9 & 29.2\\%collect a new dataset from Holl-E %bigram F1 (R-2) 23.1, R@1 evaluate knowledge selection

PIPM+KDBTS \cite{DBLP:conf/emnlp/ChenMLCXXZ20} & Holl-E* & 30.8 &   &   & & 39.2 & 30.6\\%bigram F1 (R-2) 23.9, R@1 evaluate knowledge selection

DukeNet \cite{DBLP:conf/sigir/MengRCSRTR20} & Holl-E* && --- ---(--- ---/--- ---/--- ---/19.15) & 36.53/23.02/31.46 & & & 30.3\\%meteor 30.93, evaluate knowledge selection

DiffKS(Dis) \cite{DBLP:conf/emnlp/ZhengCJH20}  & Holl-E*  &   &--- ---(--- ---/29.9 /--- ---/25.9 )&--- ---/26.4 /--- ---&  & &33.5 \\

AKGCM \cite{DBLP:journals/corr/abs-1903-10245} & Holl-E* &  & --- ---(--- ---/--- ---/--- ---/30.84) & --- ---/29.29/34.72 & & & 42.04\\% evaluate knowledge selection
%\midrule
%SEQS \cite{DBLP:conf/emnlp/ZhouPB18} & CMUDoG & & & & & 10.11 & \\
\midrule
ITDD \cite{DBLP:conf/acl/LiNMFLZ19} & DoG & & --- ---(--- ---/--- ---/--- ---/ 0.95) & & & 15.11  &\\
CAT \cite{DBLP:conf/emnlp/MaZS020} &DoG* &&--- ---(--- ---/--- ---/--- ---/ 1.22)&--- ---/--- ---/11.22&&15.2&\\%KU / QKU

DialogT \cite{DBLP:conf/ksem/TangH19} & DoG & & --- ---(--- ---/--- ---/--- ---/ 1.28) & & & 50.3 & \\

DRD \cite{DBLP:conf/iclr/ZhaoWTX0020} & DoG & 10.7 & --- ---(15.0/ 5.70/ 2.50/ 1.20) & & & 54.4  & \\% Average Extrema Greedy  0.828 0.422 0.628 

ZRKGC \cite{DBLP:journals/corr/abs-2008-12918} & DoG & 12.2 &--- ---(16.1/ 5.2 / 2.1 / 0.9 ) &&&53.8&\\
KnowledGPT \cite{DBLP:conf/emnlp/ZhaoWXTZY20}& DoG  & 13.5  &  && & 20.6 & \\
Reduce-Match \cite{DBLP:journals/access/AhnLP20} &DoG* & 13.6 & 0.7(--- ---/--- ---/--- ---/--- ---) && & 52.4 & 27.7\\%NIST 25.0% %knowledge selection 
\midrule
ZRKGC \cite{DBLP:journals/corr/abs-2008-12918} & T-Chat & 16.1& --- ---(22.3/ 8.0 / 3.7 / 1.9 )&&&42.8&\\

TF \cite{DBLP:conf/interspeech/GopalakrishnanH19} & T-Chat & 22 & & & 0.80/0.81\# & 43.6  &\\%0.22 %div-1/2  

PD-NRG \cite{DBLP:journals/corr/abs-2005-12529} &T-Chat* & 22.3&--- ---(--- ---/--- ---/--- ---/ 2.0 ) & --- ---/--- ---/10.8 & 0.022/0.181\# & 12.62 &\\%
\midrule
CMR \cite{DBLP:conf/acl/QinGBLGDCG19} & CbR & 0.38 & --- ---(--- ---/--- ---/--- ---/ 1.38) & & 0.052/0.283 &  &\\%NIST 2.238  Entropy-4 9.887 Meteor 7.46%

RAM \cite{DBLP:journals/corr/abs-2005-06128} & CbR & 0.50 & --- ---(--- ---/--- ---/--- ---/ 1.47) & & 0.053/0.287 &  &\\%NIST 2.551  Entropy-4 9.900 Meteor 7.88%

CGRG \cite{DBLP:journals/corr/abs-2005-00613} & CbR* &  & --- ---(--- ---/--- ---/--- ---/ 3.26) & &--- ---/0.116\# & &\\%NIST 1.80 %div 
\bottomrule
\end{tabular}
\end{table}

%To improve this, \citet{DBLP:conf/emnlp/MogheABK18} also employed the BiDAF which used the context as a query to select the labeled text span in a document. However, the predicted span was different from the target conversation. 

In Table \ref{HollE}, \ref{WoW} and \ref{Persona}, we summarize the auto evaluation results of generative models. For the Holl-E task in Table \ref{HollE}, the results are based on the single reference and oracle background (256 words) version of the data. We first compare some models with \textbf{implicit knowledge selection}. The GTTP \cite{DBLP:conf/acl/SeeLM17} model was first proposed for Abstractive Summarization. It was used as one of the baseline models in the original Holl-E paper \cite{DBLP:conf/emnlp/MogheABK18}, the input was a concatenation of document and dialogue context and the output was a response. The poor performance was caused by the inefficient interaction between knowledge and context. The Cake \cite{DBLP:journals/corr/abs-1906-06685} used the interaction mechanism in BiDAF \cite{DBLP:conf/emnlp/MogheABK18} to perform knowledge pre-selection then produced response by copying from knowledge or generating from the vocabulary. The RefNet \cite{DBLP:journals/corr/abs-1908-06449} extent the copy mechanism by directly selecting a semantic unit (e.g., a span containing complete semantic information) from the document. The GLKS \cite{DBLP:journals/corr/abs-1908-09528} argued that previous models adopted a local perspective (select a token based on the current decoding state). They introduced a global perspective that pre-selected some text fragments to better guide the generation. 

In the \textbf{explicit knowledge selection} models, \citet{DBLP:conf/iclr/KimAK20} focused on the role of dialogue history in knowledge selection. They introduced SKT to select knowledge of each utterance based on a latent knowledge tracking vector. Based on SKT, \citet{DBLP:conf/emnlp/ChenMLCXXZ20} enhanced the prior selection module with a Posterior Information Prediction Module (PIPM) and propose a Knowledge Distillation Based Training Strategy (KDBTS) to overcome the exposure bias in knowledge selection. \citet{DBLP:conf/sigir/MengRCSRTR20} argued that former work did not pay attention to the repetition of knowledge selection. The DukeNet they proposed had a knowledge shifter and tracker module to capture the topic flow and reduce knowledge repetition in dialogue. The DiffKS(Dis) \cite{DBLP:conf/emnlp/ZhengCJH20} model focused on the knowledge repetition problem as well. It used the dialogue context and the previously selected knowledge to compute two distributions on the candidate knowledge provided at the current turn, then combine these distributions to choose appropriate knowledge to be used in generation. Benefiting from the ability to select knowledge more accurately, DiffKS(Dis) had a higher BLEU and ROUGE than the DukeNet. The AKGCM \cite{DBLP:journals/corr/abs-1903-10245} owned the best BLEU and Hit@1, which showed reasoning on an augmented knowledge graph was more effective on knowledge selection.

For the CMUDoG task in Table \ref{HollE}, ITDD/CAT/DIalogT/DRD adopted \textbf{implicit KS} methods, while KnowledGPT and Reduce-Match used the \textbf{explicit KS} method. DRD worked on the low-resource setting, and ZRKGC worked with the zero-resource setting. Their performances were very close except that models based on GPT-2 (KnowledGPT) or a deliberation decoder (ITDD and CAT) get a lower PPL. 

The TF model is the baseline model proposed in Topical-Chat paper \cite{DBLP:conf/interspeech/GopalakrishnanH19}. The PD-NRG \cite{DBLP:journals/corr/abs-2005-12529} conducted experiments on an augmented version of Topical-Chat. They proved that dialogue action planning could benefit the response generation. Notably, the TF model computed the lexical diversity while the PD-NRG model used a corpus diversity.

For the CbR task, we compare 3 \textbf{implicit KS} models. The CMR used an MRC model to perform knowledge selection. The RAM further improved the CMR model by utilizing the response information with a Teacher-Student framework. However, the F1 on the CbR was lower than other datasets because the external knowledge was a much longer document hence the models had difficulty in locating the salient information. CGRG \cite{DBLP:journals/corr/abs-2005-00613} introduced lexical control phrases and inductive attention in a pre-trained GPT-2. It got a higher BLEU-4 than CMR and RAM that used GloVe embeddings.

%Some models (PostKS, KIC, MKST, TED) focused on learning knowledge selection ability and they did not use ground-truth label in WoW. 
%tiny scriptsize footnotesize small normalsize large Large LARGE huge Huge
\begin{table}[h]
\footnotesize
\caption{The UTED models' performance on the Test Seen version of WoW. "MT." stands for METEOR. * means different data processing schema or metric definition. "R@1" stands for Recall@1. \# means the R@1 is for response selection while others are for knowledge selection.} %\% means the column values are multiplied by 100.
\label{WoW}
\begin{tabular}{l|c|c|c|c|c|c|c}
\toprule
Model & F1\% & BLEU(BLEU-1/2/3/4)\% & ROUGE(1/2/L)\% &MT.\% & Distinct(1/2) & PPL & R@1\%\\%ROUGE (1/2/L) \% 
\midrule
MKST \cite{DBLP:conf/www/ZhaoWHYCW20}* & & 0.72(--- ---/--- ---/--- ---/--- ---) && & 0.091/0.341  & & \\%FusionF1 0.213

Adapter-Bot \cite{DBLP:journals/corr/abs-2008-12579} &18.0 &1.35(--- ---/--- ---/--- ---/--- ---) & &&&19.5 &\\%meta-knowledge 35.5 12.26 9.04 

TED \cite{DBLP:conf/cikm/ZhengZ19}* &   & --- ---(20.27/9.47/5.33/3.35) &&8.45 &0.039/0.162&  & \\%Meteor 8.45%. seen/unseen unclear

DRD \cite{DBLP:conf/iclr/ZhaoWTX0020}* & 18.0 & --- ---(21.8 /11.5 / 7.5 / 5.5 )  &&  & & 52.0 & \\%Average 83.5 Extrema 42.7 Greedy 65.8

ZRKGC \cite{DBLP:journals/corr/abs-2008-12918}* & 18.9 &--- ---(22.5 / 8.4 / 3.9 / 2.0 ) && & &41.1&\\

KIC \cite{DBLP:conf/acl/LinJHWC20}* & &  --- ---(17.3 /10.5 / 7.7 /--- ---) &&& 0.138/0.363 & &\\ 
%KIF \cite{DBLP:journals/corr/abs-2004-12744}  & WoW & 26.9 &&&&&\\%two sources of external information

Decoupling \cite{DBLP:journals/corr/abs-2004-14614}* & 20.1 &&&& & 18.3  & 90.84\# \\%selection most related article as knowledge,The selection is from 100 candidates in WOWs, and 20 candidates in other datasets. 

Unlikelihood \cite{DBLP:conf/acl/LiRKWBCW20} & 35.8 &&&&& 8.5 &\\ %standard perplexity and F1 metrics. 0.358
\midrule
PostKS \cite{DBLP:conf/ijcai/LianXWPW19}* & 1.74 &--- ---(17.2 / 6.9 / 3.4 /--- ---)&&  &0.056/0.213&&\\
Match-Reduce \cite{DBLP:journals/access/AhnLP20}  & 17.8 & 1.2 (--- ---/--- ---/--- ---/--- ---) &&& & 60.6 & 25.4\\%NIST 46.7%
TMN \cite{DBLP:journals/corr/abs-1811-01241} & 18.9 &   & & & & 46.5 & \\
DiffKS(Dis) \cite{DBLP:conf/emnlp/ZhengCJH20}* &   &--- ---(--- ---/11.3 /--- ---/ 5.7 )&--- ---/6.8 /--- ---&  & & &24.7 \\
DukeNet \cite{DBLP:conf/sigir/MengRCSRTR20}*  & & --- ---(--- ---/--- ---/--- ---/2.43) &25.17/6.81/18.52 &17.09*  & & & 26.38\\%meteor 17.09 rouge 25.17/6.81/18.52 
SKT \cite{DBLP:conf/iclr/KimAK20}*  & 19.3  &   &  && & 52.0 & 26.8\\ %bigram F1 (R-2) 6.8,  the accuracy for knowledge selection

PIPM+KDBTS \cite{DBLP:conf/emnlp/ChenMLCXXZ20}* & 19.9  &   &  && & 42.7 & 27.9\\ %bigram F2 (R-2) 7.3,  the accuracy for knowledge selection

AKGCM \cite{DBLP:journals/corr/abs-1903-10245}  &  & --- ---(--- ---/--- ---/--- ---/6.94) &--- ---/7.38/17.02&  & & & 18.24\\%ROUGE --- ---/7.38/17.02

KnowledGPT \cite{DBLP:conf/emnlp/ZhaoWXTZY20}& 22.0  &   &  && & 19.2 & \\

Interleave \cite{DBLP:journals/corr/abs-1911-09728} * & 35.7 &&&&&19.7&\\%seen/unseen unclear

\bottomrule
\end{tabular}
\end{table}

In Table \ref{WoW}, we introduce the models' performance on the WoW dataset. In the \textbf{implicit KS} models (the first half), different data processing schemes are usually adopted. MKST \cite{DBLP:conf/www/ZhaoWHYCW20} and Adapter-Bot\footnote{Adapter-Bot powered with meta-knowledge has a big improvement (F1 35.5, BLEU 12.26, PPL 9.04).} \cite{DBLP:journals/corr/abs-2008-12579} proposed methods to utilize different types of knowledge. MKST employed Pre-trained Transformer while Adapter-Bot adopted DialoGPT for multi-task learning and used BERT as a task identification. TED \cite{DBLP:conf/cikm/ZhengZ19} performed KS with attention in the decoder. DRD \cite{DBLP:conf/iclr/ZhaoWTX0020} and ZRKGC \cite{DBLP:journals/corr/abs-2008-12918} both utilized PTMs and got competitive results in limited-resource setting. KIC \cite{DBLP:conf/acl/LinJHWC20} designed a Knowledge-Interaction and Knowledge-Copy mechanism. The better distinction value indicated that copying from both knowledge and context could produce a more diverse response. \citet{DBLP:journals/corr/abs-2004-14614} trained the Decoupling model to automatically recouple context and related knowledge, their model performance on response selection is impressive. Unlikelihood \cite{DBLP:conf/acl/LiRKWBCW20} designed 4 novel losses to address the flaws in response generation, such as repetition and logic. The best F1 and PPL results showed these loss functions could facilitate model convergence. 

In the \textbf{explicit KS} models (the second half), PostKS \cite{DBLP:conf/ijcai/LianXWPW19} employed a Teacher-Student module to leverage the response information. They employed a seq-to-seq model without a copy mechanism and did not use the ground-truth knowledge label. This caused their model hard to train and obtained the lowest F1 compared with other models. Match-Reduce \cite{DBLP:journals/access/AhnLP20} model matched each context utterance with knowledge sentences to capture fine-grained interactions and aggregated them as a training loss. TMN \cite{DBLP:journals/corr/abs-1811-01241} was a baseline model that used a Transformer Memory network with explicit KS. The DiffKS(Dis) \cite{DBLP:conf/emnlp/ZhengCJH20}, AKGCM \cite{DBLP:journals/corr/abs-1903-10245}, DukeNet \cite{DBLP:conf/sigir/MengRCSRTR20}, SKT \cite{DBLP:conf/iclr/KimAK20} and PIPM+KDBTS \cite{DBLP:conf/emnlp/ChenMLCXXZ20} conducted experiments on both Holl-E and WoW. The Holl-E dataset is a relatively simpler dataset compared with WoW. DiffKS(Dis) and AKGCM \cite{DBLP:journals/corr/abs-1903-10245} performed KS better than the other three models on Holl-E but performed worse than them on WoW. The reason might be the different encoders they employed. DukeNet, SKT, and PIPM+KDBTS employed a BERT encoder which might encounter the over-fitting problem on Holl-E. This caused their performance worse than the relatively simple encoding method (Glove+RNN) DiffKS and AKGCM employed. Notably, AKGCM got a 42.04\% Recall@1 on Holl-E, around 10\% higher than the other 4 models. While on WoW, AKGCM only had an 18.24\% Recall@1, around 9\% lower than the other 4 models. Besides the difference of datasets, the convergence instability of RL they used when performing KS might lead to unstable performance. DukeNet employed METEOR Universal \cite{DBLP:conf/wmt/DenkowskiL14}, a language-specific version of the METEOR metric. KnowledGPT \cite{DBLP:conf/emnlp/ZhaoWXTZY20} had the best PPL showing the effectiveness of GPT-2 \cite{radford2019language}. Interleave \cite{DBLP:journals/corr/abs-1911-09728} used different decoder layers to attend knowledge or context in a Seq-to-Seq Transformer. They had the highest F1 among the explicit KS models.

Among these models, only Match-Reduce, TED, KIC, and Interleave randomly initialized word embeddings. Generally speaking, their performances are slightly lower than models with pre-trained embeddings except Interleave because of different data processing schemes.

\begin{table}[h]
\footnotesize
\caption{The generative models' performance on Persona-Chat/ConvAI2/DSTC-7/ED/LIGHT datasets. "Persona" is short for Persona-Chat. "MT." stands for METEOR. * means a difference in data processing or metric definition. "Ent-4" is short for Entropy-4. "R1" stands for Recall@1 and is for response selection.}%\% means the column values are multiplied by 100.\# means the R@1 is for candidate selection while others are for knowledge selection.
\label{Persona}
\begin{tabular}{l|c|c|c|c|c|c|c|c}
\toprule
Model & Dataset & F1\% & BLEU(BLEU-1/2/3/4)\% & MT.\% &Ent-4 & Distinct(1/2) & PPL & R1\%\\%ROUGE (1/2/L) \% 
\midrule
Profile Memory \cite{DBLP:conf/acl/KielaWZDUS18} &Persona &&&&&&34.5&12.5\\ %Hits@1 0.125, The latter consists of choosing 19 random distractor response
%RR \cite{DBLP:conf/emnlp/WestonDM18} & Persona & &&&&9.2&\\

RCDG \cite{DBLP:conf/aaai/SongZH020} & Persona & & && &--- ---/0.127 &29.9 & \\%Average:66.9 Greedy:47.2 Extrema:46.8		
PostKS \cite{DBLP:conf/ijcai/LianXWPW19} & Persona* &8.7&--- ---(19.0 / 9.8 / 5.9 /--- ---)& &  &0.046/0.134&&\\
GDR \cite{DBLP:conf/acl/SongWZLL20} & Persona & & && & 0.037/0.227  & 16.7&\\%Entail diin/bert  21.5%/13.0%

%\midrule
%Per-CVAE \cite{DBLP:conf/ijcai/SongZCWL19} &&&&&0.03/0.22\\
%\midrule
%COMPAC \cite{DBLP:journals/corr/abs-2010-03205} & Persona* &&--- ---(4.12/1.82/--- ---/--- ---)  &&0.0087/0.0107&16.21&\\%%%%%%?????

PAML \cite{DBLP:conf/acl/MadottoLWF19} &Persona* && 0.74(--- ---/--- ---/--- ---/--- ---)&&&&41.6& \\%C 0.20

CMAML \cite{DBLP:conf/acl/SongLBYZ20} &Persona* && 1.70(--- ---/--- ---/--- ---/--- ---)&&&0.021/--- ---&36.3&\\%C 0.18

MKST \cite{DBLP:conf/www/ZhaoWHYCW20} & Persona & & 0.82(--- ---/--- ---/--- ---/--- ---) & && 0.095/0.361  & & \\%FusionF1 0.392

Adapter-Bot \cite{DBLP:journals/corr/abs-2008-12579}   & Persona & 15.0 & 1.55(--- ---/--- ---/--- ---/--- ---) && &&11.1 &\\
Multi-Input \cite{DBLP:conf/acl/GolovanovKNTTW19} & Persona* & & 2.77(--- ---/--- ---/--- ---/--- ---)&7.88&9.211 &--- ---/0.155 &&\\%Meteor 0.07878  NIST 1.278 Entropy-4 9.211 
TF+AMI \cite{DBLP:journals/corr/abs-2007-00067} &Persona & &&&9.695& 0.104/0.385&&\\ %Average Greedy Extrema 0.869 0.672 0.433 Entropy-4 9.695

L-PCFG \cite{DBLP:conf/naacl/DuB19} &Persona* & &--- ---(20.9*/--- ---/--- ---/--- ---) & 7.77 &&&\\%Meteor 0.0777 rouge --- ---/--- ---/0.1565

PEE \cite{DBLP:conf/ecai/XuLYRRC020} &Persona& 18.4 &--- ---(23.19/11.52/6.12/3.50)&&&&&\\%Average Extrema Greedy 0.8691 0.5010 0.6906

$\mathcal{P}^2$ Bot \cite{DBLP:conf/acl/LiuCCLCZZ20} &Persona& 19.8 & &&&&15.1&81.9\\ % Given 20 response candidates, Hits@1 is the probability that the real response ranks the highest according to the model. 
Decoupling \cite{DBLP:journals/corr/abs-2004-14614}     & Persona & 19.9 &&& && 18.7  & 66.67 \\

%%%%%%Explicit
GE \cite{DBLP:conf/acl/BaoHWLW19} &Persona* & 20.5 &&&&0.021/0.097&&\\%give a WoW F1 definition 0.2050

\midrule
Deepcopy \cite{DBLP:conf/sigdial/YavuzRCH19} &ConvAI2&& 4.09(--- ---/--- ---/--- ---/--- ---)&&&--- ---/0.059&54.6&\\%Rouge --- ---/--- ---/60.30

LIC \cite{golovanov2020lost} & ConvAI2 & 17.7 &&&&&&17.1\\%accuracy of retrieving the ground truth next utterance among 19 random distractor responses sampled from other dialogues 

Unlikelihood \cite{DBLP:conf/acl/LiRKWBCW20} & ConvAI2 & 19.3 &&&&& 11.9 &\\%0.193

TransferTransfo \cite{DBLP:journals/corr/abs-1901-08149} &ConvAI2 &19.5 &&&&&16.3&80.7\\%the accuracy of retrieving a gold next utterance among 19 random distractor responses sampled from other dialogues

BART \cite{DBLP:conf/acl/LewisLGGMLSZ20} & ConvAI2 & 20.7 &  &&& &11.9&\\

\midrule
CG+CVAE \cite{DBLP:journals/corr/abs-1903-09813} & DSTC-7* & & --- ---( 3.90/0.89/0.22/0.07) &2.62&6.427&0.012/0.027 &\\%3.898/0.817/0.223/0.074 NIST-1/2 0.023/0.024  Meteor 2.62 Entropy-4 6.427

TED \cite{DBLP:conf/cikm/ZhengZ19} & DSTC-7* &   & --- ---(14.18/5.01/2.11/0.95) &5.60&& 0.022/0.094&  &\\%Meteor 5.6%. processed DSTC

CBS \cite{DBLP:journals/csl/Tam20} & DSTC-7 && --- ---(--- ---/--- ---/--- ---/ 1.83) &8.07&9.030&0.109/0.325 &\\%NIST-4 2.523 Meteor 8.07% Entropy-4 9.030 

CDD \cite{DBLP:journals/corr/abs-2005-03174} & DSTC-7 & &  --- ---(--- ---/--- ---/--- ---/ 2.43) && &0.109/0.442 & &\\% PMI 
\midrule
Adapter-Bot \cite{DBLP:journals/corr/abs-2008-12579}   & ED &   & 8.53(--- ---/--- ---/--- ---/--- ---) && &&12.2 &\\
\midrule
Decoupling \cite{DBLP:journals/corr/abs-2004-14614}     & LIGHT & 18.7 &&&& & 23.2  & 59.63 \\
\bottomrule
\end{tabular}
\end{table}

%models comparable in one paragraph
In Table \ref{Persona}, the ConvAI2 competition was primarily based on the Persona-Chat dataset and there was a slight difference in data size between them. For the Persona-Chat task, we find that most models applied the \textbf{implicit KS} except GE \cite{DBLP:conf/acl/BaoHWLW19}, a Generation-Evaluation framework that adopted \textbf{explicit KS} method. 

Profile Memory \cite{DBLP:conf/acl/KielaWZDUS18} was the baseline model proposed in the original paper. It employed a key-value memory network for persona selection. RCDG \cite{DBLP:conf/aaai/SongZH020} employed RL and NLI techniques to generate persona consistent dialogues. GDR \cite{DBLP:conf/acl/SongWZLL20} was a three stages method (generate-delete-rewrite) that deleted inconsistent terms from a generated response and further rewrote it to a context-consistent one. The better Distinct-2 and PPL results than RCDG showed the method's advantage. 

%models with specific method in one paragraph
PAML \cite{DBLP:conf/acl/MadottoLWF19} utilized meta-learning to learn better initial parameters which could be quickly adapted to new personas by leveraging only a few dialogue samples collected from the same user. CMAML \cite{DBLP:conf/acl/SongLBYZ20} proposed an algorithm based on meta-learning that customized a unique dialogue model for each task in the few-shot setting. MKST \cite{DBLP:conf/www/ZhaoWHYCW20} and Adapter-Bot \cite{DBLP:journals/corr/abs-2008-12579} aimed to leverage diverse knowledge source. The former had the second-best Distinct and the latter got the best PPL. The same trends are observed on WoW and ED tasks. It indicated that the utilization of multiple types of knowledge benefited the UTED task. Multi-Input \cite{DBLP:conf/acl/GolovanovKNTTW19} employed a PTM (GPT-1) to compare a number of architectures and training schemes. TF+AMI \cite{DBLP:journals/corr/abs-2007-00067} introduced Adversarial Mutual Information (AMI) to identify joint interactions between source and target. Since the model used randomly initialized embedding on an LSTM encoder, it was the Adversarial training object that helped their model to get the best Distinct and Entropy-4 value among all models. 

\citet{DBLP:conf/naacl/DuB19} generated the words of a sentence according to the order of their first appearance in its lexicalized PCFG (L-PCFG) parse tree instead of the traditional left-to-right manner. PEE \cite{DBLP:conf/ecai/XuLYRRC020} was a neural topical expansion framework, which was able to extend the predefined knowledge with semantically correlated content before utilizing them to generate dialogue responses. The BLEU-1/2/3 results were the best, showing some advantages of the knowledge expansion method. Decoupling \cite{DBLP:journals/corr/abs-2004-14614} and $\mathcal{P}^2$ Bot \cite{DBLP:conf/acl/LiuCCLCZZ20} were both based on PTM (GPT-1). The former focused on latent knowledge selection and the latter paid attention to model the understanding scheme between interlocutors. They both performed well on the R@1 metric, indicating that these methods were sensitive to differences between knowledge sentences. 

For the ConvAI2 task, Deepcopy \cite{DBLP:conf/sigdial/YavuzRCH19} had a better BLEU than all models. However, its Distinct-2 was not good. This was contrary to the model's motivation of using copy mechanism to enhance the ability of knowledge utilization. It showed that the auto evaluation metircs based on word overlap were insufficient to reflect the dialogue quality. LIC \cite{golovanov2020lost} and TransferTransfo\footnote{\citet{DBLP:conf/emnlp/KimKK20} proposed a self-consciousness method and a distractor memory to improve the persona consistency in dialogue. They tested their methods with three models including TransferTransfo.} \cite{DBLP:journals/corr/abs-1901-08149} used PTM (GPT-1) and fine-tuned on ConvAI2. Unlikelihood \cite{DBLP:conf/acl/LiRKWBCW20} method had an competitive performance compared with BART \cite{DBLP:conf/acl/LewisLGGMLSZ20} which was a PTM with denoising autoencoder and conducted experiments on ConvAI2 task. Among the results of Persona-Chat and ConvAI2, we observed that models with PTMs normally had a better PPL than others.

The DSTC-7 task was based on the CbR task except that the external knowledge was organized as sentences, not documents. CG+CVAE \cite{DBLP:journals/corr/abs-1903-09813} tried different methods to calculate the latent variable in CVAE, but the GRU-based encoder may restrict the performance. TED \cite{DBLP:conf/cikm/ZhengZ19} improved the performance with a Transformer-based model with randomly initialized embeddings. CBS \cite{DBLP:journals/csl/Tam20} introduced Cluster-based Beam Search and exceeded TED in all metrics with a simple LSTM+copy framework. CDD \cite{DBLP:journals/corr/abs-2005-03174} had a switcher to control whether to perform convergent or divergent decoding, it outperformed the CBS in BLEU-4 and Distinct-2. Since these models employed the random initialized embeddings or GloVe embeddings on simple encoders, we reckon that models based on large parameter PTMs will achieve better performance on this task.

%Adapter-Bot \cite{DBLP:journals/corr/abs-2008-12579} was also tested on ED and Decoupling \cite{DBLP:journals/corr/abs-2004-14614} was also experimented on LIGHT.  

\subsection{Summary of experimental results }

From the current UTED experimental results, we find some interesting conclusions:
1) Adding a copy mechanism is always helpful in Generative models \cite{DBLP:journals/corr/abs-1908-06449,DBLP:journals/corr/abs-1908-09528,DBLP:conf/iclr/KimAK20,DBLP:conf/sigir/MengRCSRTR20};
2) PTMs are better at leveraging semantic information and can improve the models' performance in most cases; 
3) Meta-learning is now only applied to the Persona-chat dataset and does not show obvious advantages. How to leverage the meta-learning in UTED requires more investigation;
4) Supplementary information is useful, such as knowledge expansion \cite{DBLP:journals/corr/abs-2010-03205}, syntactic tree structures\cite{DBLP:conf/naacl/DuB19}, meta-knowledge \cite{DBLP:journals/corr/abs-2008-12579}; 5) Experimental results on most datasets in retrieval models can still improve since the R@1 is below 83\% and the candidate sizes are small; 
6) By observing the range of various metrics in generative models: BLEU in (0.7 to 8.6)\%, METEOR in (2.6 to 8.5)\%, Distinct-2 in (0.025 to 0.445), F1 in (1.7 to 36)\%, we can say that despite many different structures and algorithms that researchers have applied, the dialogue quality and knowledge utilization of the UTEDS still have a lot to improve.

%====================Section 7: Future trends===================
\section{Future Trends}
\begin{figure}[h]
\centering
\includegraphics[width=4.5in]{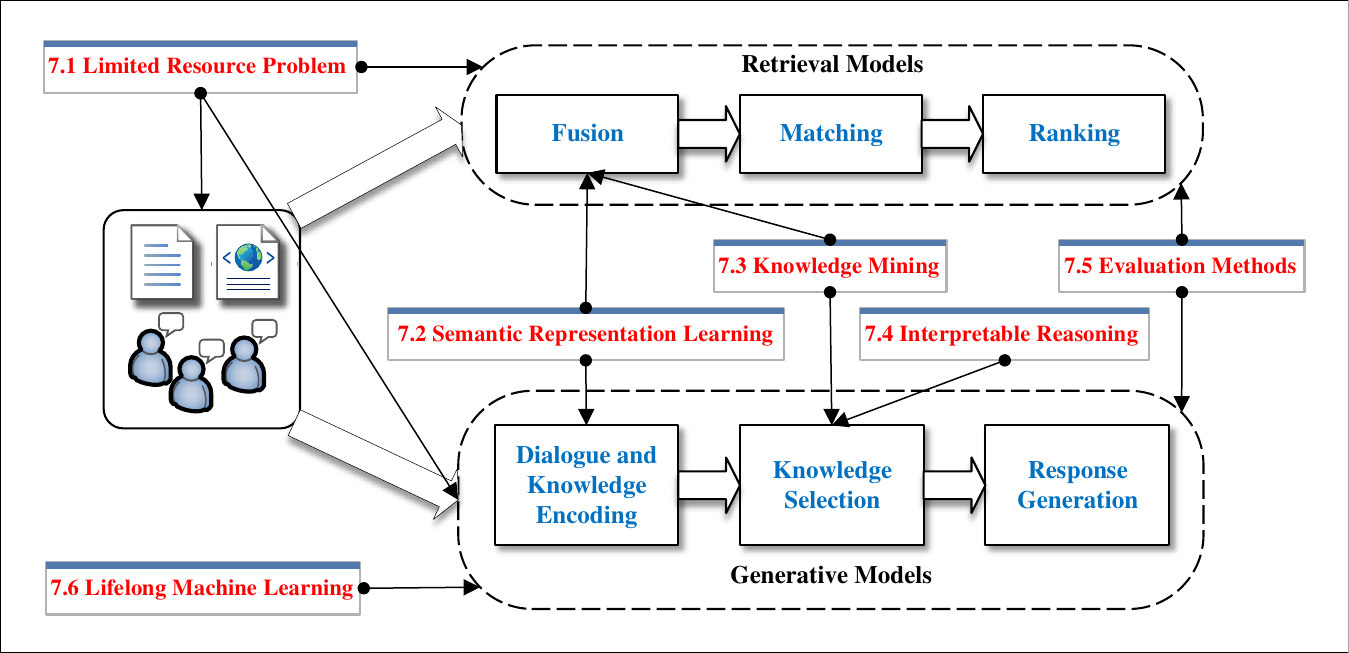}
\caption{The future research trends of the UTEDS. Each direction points to the related modules of Retrieval and Generative models.}
\label{UTEDS-FT}
\end{figure}

So far we introduced UTEDS from the perspective of retrieval and generative. The retrieval models can choose fluent responses from given-candidates but have difficulty adapting to specific dialogue context and external knowledge. In contrast, generative models can make better use of context and external knowledge and attract more attention from researchers. Benefiting from the current powerful PTMs \cite{radford2019language,DBLP:conf/acl/ZhangSGCBGGLD20}, generative models have made significant improvements in producing fluent responses but still have room for improvement. Figure \ref{UTEDS-FT} shows 6 future research trends for UTEDS. 7.1/7.2/7.3/7.5 refer to both retrieval and generative models, 7.4/7.6 are only suitable for generative models.

\subsection{Limited Resource Problem} 
%%%%%%Despite the ubiquitous text resource in internet, the UTED dataset nowadays still have some problem.
The UTED aims to imitate the real human dialogues where interlocutors can freely refer to their own knowledge. However, the current dataset constructed by researchers has the following problems: 1) \textbf{Small data size.} Some datasets (ARW \cite{DBLP:conf/coling/VougiouklisHS16}, CMU\_DoG \cite{DBLP:conf/emnlp/ZhouPB18}, Holl-E \cite{DBLP:conf/emnlp/MogheABK18}, etc.) are not big enough for training large parameter models \cite{DBLP:conf/emnlp/ZhouPB18}; 2) \textbf{Knowledge sparsity.} Some datasets (GCD \cite{DBLP:conf/aaai/GhazvininejadBC18}, CMU\_DoG \cite{DBLP:conf/emnlp/ZhouPB18}, CbR \cite{DBLP:conf/acl/QinGBLGDCG19}, etc.) suffer from knowledge sparsity \cite{DBLP:conf/aaai/ZhengZHM20,DBLP:journals/corr/abs-2004-02644} problem which means the dialogue data collected only contains little external knowledge. This type of data distribution makes it difficult for the model to learn to inject knowledge into conversation.

% The only dataset which do not suffer this knowledge sparsity is Holl-E. When giving a response to a dialogue context, the workers copy the existing fragment in the background document, only add a few words in front or back to make the response more fluent to the context. However, this data construction strategy is still far from the real dialogue scenario, because the knowledge referred by speakers in the dialogues usually does not exactly match the source. Besides, the Holl-E also suffers the small data size problem.

The limited resource problem can be addressed in two directions. The first is \textit{data construction}, a large-scale dataset with clear knowledge utilization will be helpful to train the UTED model. Due to the expense of artificial data construction, we can focus on large-scale dialogue data that the knowledge utilization is naturally labeled (e.g. when consulting legal texts, the legal knowledge involved in the dialogue has a clear source and is easy to locate.). The second is \textit{model structure}, a specific architecture focusing on knowledge sparsity can be designed. In a broad sense, we consider the human dialogues to consist of linguistic knowledge, dialogue intentions, and world knowledge. Linguistic knowledge is the grammatical and syntactical rules, which can be learned by PTMs \cite{DBLP:journals/corr/abs-2003-08271}. The dialogue intentions is "what to act" in a dialogue (e.g. strategies in persuasion \cite{DBLP:conf/acl/WangSKOYZY19}, negotiation \cite{DBLP:conf/sigdial/ZhouHBT19} or recommendation \cite{DBLP:conf/emnlp/HayatiKZSY20} tasks, yes-and dialogue strategy \cite{DBLP:conf/acl/ChoM20}). The world knowledge is "what to refer" in a dialogue, including commonsense knowledge, entities, and events, etc. The problem is, some of the dialogue intentions require world knowledge injection, others do not. Even among the dialogue intentions which need knowledge injection, there is a difference between the knowledge utilization pattern. These differences cause the UTED model hard to learn the knowledge injection in different dialogue intentions. Therefore, we believe that a good UTED model needs to set up different modules to identify the dialogue intentions and control the world knowledge injection.

Besides the above two problems, there are \textbf{other limited resource problems}: 1) the amount of knowledge available in different languages is extremely imbalanced. For instance, the number of articles in English is hundreds of times larger than that of Bengali on Wikipedia. This limited resource problem can be addressed with cross-lingual \cite{DBLP:conf/aaai/Chi0WWMH20,DBLP:conf/acl/ConneauKGCWGGOZ20} methods. Retrieving relevant text from external knowledge in a rich resource language can improve the informativeness of the response generation in the low resource language; 2) many specialized domains contain their own specific terms that are not part of the pre-trained LM vocabulary. Some cross-domain \cite{DBLP:conf/emnlp/ZhangRSCFFKRSW20} methods can be used to address this problem. Researchers can tap more potential from the cross-lingual and cross-domain methods.
%should the third problem put together with the first two?

%An intuitive impression is that the dialogue intentions is limited and the world knowledge is infinite. However, in current UTED dataset, the text related to the dialogue intentions is much more than text related to world knowledge. This bias

%We outline two possible directions for future research:  1) \textbf{Proper pre-training data.} UTED \cite{DBLP:conf/nips/00040WWLWGZH19,DBLP:conf/ijcai/XiaoZLST0W20,DBLP:conf/acl/LewisLGGMLSZ20} can benefit from PTM. However, a larger amount of pre-training data do not necessarily  better results \cite{DBLP:conf/acl/GururanganMSLBD20}. The use of proper domain or task data for pre-training can achieve the best performance. Identify and measure the most useful data distribution for the pre-training corpus is an important research aspect. This could also benefit the domain transfer problem. 2)\textbf{Effective Representation learning.} 

\subsection{Semantic Representation Learning}
Semantic representation learning aims to learn universal language representations that contain salient language knowledge, which can be understood and used by a computer. It is a research Hotspot in NLP \cite{DBLP:conf/naacl/PetersNIGCLZ18,DBLP:conf/naacl/DevlinCLT19,radford2018improving,radford2019language}. 

In UTED, data distribution is different between knowledge and dialogue, and between different types of knowledge. For example, the language used in movie reviews is different from movie plots. We need the UTEDS that can better integrate dialogue and knowledge, understand the relationship between dialogue and knowledge, identify, select, and use knowledge according to the dialogue. The current knowledge representation methods are not up to the requirements. Some possible directions are: 1) borrowing ideas from the latest PTMs to fully utilize the structured representation ability of hidden variable space \cite{DBLP:conf/emnlp/LiGLPLZG20,DBLP:conf/iclr/DathathriMLHFMY20}. The latent spaces can be used to represent knowledge with different data distribution. 2) the current models usually adopted PTMs and fine-tuning to distinguish tasks from the parameter perspective but ignored the model-structure perspective, resulting in similar dialogue models for different tasks. We can design specific model architecture to jointly learn dialogue and knowledge representations \cite{DBLP:journals/corr/abs-2002-08909}. 3) using task-related domain data to re-train PTM \cite{DBLP:conf/acl/GururanganMSLBD20}. %For example, we can mask some knowledge related words.

\subsection{Knowledge Mining}
We present 2 knowledge mining categories: 1) \textbf{Latent dialogue knowledge.} The UTEDS need to mine information not only from external knowledge but also from dialogue context. Latent attributes of interlocutors can help to build consistent and engaging DS. \citet{DBLP:conf/www/TigunovaYMW19} addressed this type of knowledge acquisition problem by extracting personal attributes from the conversation. Specifically, user-generated social media text (Reddit) is used to infer the user’s experience from voice conversations (movies and Persona-Chat). The current research on extracting dialogue knowledge focused on exploiting persona \cite{DBLP:conf/acl/BoydPSPC20}, emotion \cite{DBLP:conf/emnlp/LinMSXF19}, or action \cite{DBLP:conf/acl/BaoHWWW20}. Future research can explore more user behaviors, habits, and other latent characteristics. 2) \textbf{Long text knowledge selection.} Unstructured knowledge usually comes from web-page text (Wikipedia, news, etc.), and these web page texts usually contain thousands of words. The existing methods for processing long texts used MRC technology to implicitly filter the entire text or explicitly truncate the text into independent sentences. The performance of implicit filtering decreases significantly as the length of the text increases\footnote{For example, on the Holl-E dataset, the RefNet model got 29.38 and 17.19 BLEU when using 256 and 1200 words external knowledge, respectively.}, while explicit filtering ignores the logical relationship between sentences and wastes text information. A possible direction is to integrate the advantages of implicit and explicit methods for long text knowledge selection. %For example, integrating the external knowledge retrieval into pre-training \cite{DBLP:journals/corr/abs-2002-08909}, learning to \cite{}

\subsection{Interpretable Reasoning}
Interpretable reasoning means that the model can give an explicit, logical and complete reasoning path to explain the reason of choosing the corresponding knowledge. Interpretable reasoning can not only help understand machine's reasoning logic but also meet the ethical requirements of artificial intelligence. For a long time, researchers have shown a great interest in constructing models with specific reasoning abilities, such as algebraic reasoning \cite{DBLP:conf/aaai/Clark15}, logical reasoning \cite{DBLP:journals/corr/WestonBCM15}, commonsense reasoning \cite{DBLP:conf/semeval/OstermannRMTP18}, multi-fact reasoning \cite{DBLP:conf/aclnut/WelblLG17,DBLP:conf/kdd/ShenHGC17,DBLP:conf/naacl/TalmorB18,DBLP:conf/naacl/DuaWDSS019}, and multiple steps reasoning \cite{DBLP:conf/naacl/KhashabiCRUR18,DBLP:conf/emnlp/MihaylovCKS18,DBLP:journals/tacl/WelblSR18,DBLP:conf/emnlp/Yang0ZBCSM18}. 

In UTED, most models depended on the attention mechanism to conduct interpretable reasoning, but the effectiveness of attention for interpretability is still controversial \cite{DBLP:conf/naacl/JainW19,DBLP:conf/acl/SerranoS19}. To address this, some work first used unstructured knowledge to construct a concrete memory (augmented knowledge graph \cite{DBLP:journals/corr/abs-1903-10245} or event graph \cite{DBLP:conf/ijcai/XuLWN0C20}), and then adopted reinforcement learning to reason on the memory. These explicit KS methods had difficulty in leveraging paragraph-level or document-level text information compared with attention-based methods. To conduct interpretable reasoning and leverage the global text information, we believe a more fine-grained structure that using both implicit KS in text and explicit KS in structured data is the future trend. Meanwhile, some traditional methods which possess good interpretability, such as symbolic logical reasoning, can be leveraged.

\subsection{Evaluation Methods}
Generally speaking, DS with good performance needs to produce a response with high semantic relevance, rich information, and diverse expressions. In UTEDS, the evaluation needs to reflect not only the dialogue quality but also the external knowledge utilization. For example, \citet{DBLP:conf/www/ZhaoWHYCW20} defined the Fusion F1 (F1 \cite{DBLP:journals/corr/abs-1811-01241}+Knowledge F1 \cite{DBLP:conf/ijcai/LianXWPW19}) as an automatic metric to evaluate the quality of responses. While F1 evaluates the char-based F-score of prediction against gold response, Knowledge F1 evaluates the exact recall performance of the output response at the char level relative to the text knowledge. However, due to the expense of human evaluation and the deficiency of auto evaluation metrics in dialogue scenario \cite{DBLP:conf/acl/LoweNSABP17,DBLP:conf/aaai/TaoMZY18}, we still need to find more reliable auto evaluation metrics highly correlated with human evaluation. %explain drawbacks

We believe that model-based evaluations are the future direction. Although earlier methods such as ADEM \cite{DBLP:conf/acl/LoweNSABP17} which mimic human evaluation have been showed insensitive in adversarial scenarios \cite{DBLP:conf/aaai/SaiGKS19}, recent research \cite{DBLP:conf/cikm/BolotovaBZCSS20} outlined that the current PTM can effectively approximate human annotations when annotating paragraph-level text. Another recently proposed USR \cite{DBLP:conf/acl/MehriE20} model leveraged the ability of PTM and used a regression model to approximate the specific scores rated by annotators. These studies shed light on the future direction of auto evaluations. However, there are still difficulties in training evaluation models because of the human-annotated training dataset: 1) the definition of human evaluation metrics lacks a unified standard. For example, there are no universal standards for what is an interesting response; 2) the workers with different knowledge backgrounds tend to give very different scores on the same metric; 3) the annotation data size is small since the human annotations are expensive. We need to solve these problems which cause bias accumulation in training.

%Other than what we introduced in subsection 6.1, there are many works that addressed the auto evaluation methods of the DS. \citet{DBLP:conf/sigdial/Paek01} studied what purpose dialogue measurement serves, and then propose an empirical method to evaluate the system that meets that purpose. \citet{DBLP:journals/corr/KannanV17} discussed the adversarial evaluation in the DS. \citet{DBLP:conf/naacl/XuJLRWWWW18} proposed mean diversity score (MDS) and probabilistic diversity score (PDS) to evaluate the diversity of responses generated when multiple reference responses are given. \citet{DBLP:conf/aaai/TaoMZY18} evaluated a reply by taking into consideration both a ground-truth reply and a query. DiscoveryScore \cite{DBLP:conf/conll/ZemlyanskiyS18} can be used as automatic metrics for dialogues quality since it is strongly correlated with Engagingness score. \citet{DBLP:conf/aaai/SaiGKS19} pointed out the drawback of the ADEM and outlined we still have a long way to go in the automatic evaluation of DS. \cite{DBLP:conf/naacl/DziriKMZ19} presented interpretable metrics for evaluating topic coherence by making use of distributed sentence representations. \citet{DBLP:conf/sigdial/MehriE20} proposed fine-grained evaluation of dialogue (FED), an automatic evaluation metric which used DialoGPT \cite{DBLP:conf/acl/ZhangSGCBGGLD20}, without any fine-tuning or supervision. They also introduced a dataset with 18 different dialogue quality labels in turn-level and dialogue-level.

\subsection{Lifelong Machine Learning}
Lifelong machine learning (LML) \cite{DBLP:journals/ras/Steels95,DBLP:conf/nips/Thrun95,DBLP:conf/icml/Chen014,DBLP:conf/aaai/MitchellCHTBCMG15} requires that the deployed machine learning system have the capability to continuously improve themselves through interaction with the environment. As we introduced in Table \ref{diff_of_datasets}, the UTED datasets involve different domains. We hope that a UTEDS has a certain memory ability and can learn a new domain knowledge without forgetting the old one, which is called \textbf{knowledge retention} \cite{DBLP:conf/icml/Chen014}. We also hope that a UTEDS can comprehend by analogy, and can perform well in the new domain through existing knowledge, which is called \textbf{knowledge transfer} \cite{DBLP:journals/ras/Steels95}. 

Knowledge retention and knowledge transfer can be seen as high-level abstract memory ability. The UTEDS with lifelong learning ability needs to design a memory module with appropriate spatial structure, distinguish good experience from bad ones, preserve experience knowledge, slow down catastrophic forgetting problem, update the obsolete knowledge, establish new knowledge in the face of unseen tasks, and balance the mutual influence of memory ability and performance. There is a lack of architecture, dataset, and benchmarks to test this high-level memory ability of the UTEDS.

%%=====================Conclusion================
\section{Conclusion}
The UTEDS need to select correct external knowledge according to dialogue and incorporate the knowledge into response generation. We believe that extracting unstructured text information during dialogue is the future trend in DS research because a large amount of human knowledge is contained in these texts. The research of the UTEDS not only possesses a broad application prospect but also facilitates the DS to better understand human knowledge and natural language. This article introduces the UTEDS, defines the related concepts, analyzes the current datasets/model structures/evaluation methods/model performance, and provides views on future research trends, hoping to help researchers in this field.

\section{Acknowledgments}
This paper is supported by the National Natural Science Foundation of China (No. 62076081, No. 61772153 and No. 61936010) and Science and Technology Innovation 2030 Major Project of China (No. 2020AAA0108605) and Nature Scientific Foundation of Heilongjiang Province(YQ2021F006).

\bibliographystyle{ACM-Reference-Format}
\bibliography{sample-base}

\end{document}